\RequirePackage{fix-cm}
\documentclass[twocolumn]{svjour3}
\smartqed

\usepackage{bm}
\usepackage{graphicx}%
\usepackage{multirow}%
\usepackage{newtxtext, newtxmath}
\usepackage{natbib}
\usepackage{balance}
\usepackage[colorlinks=true,citecolor=blue]{hyperref}
\usepackage{amsmath,amsfonts}%
\usepackage{mathrsfs}%
\usepackage[title]{appendix}%
\usepackage{xcolor}%
\usepackage{textcomp}%
\usepackage{manyfoot}%
\usepackage{booktabs}%
\usepackage{algorithm}%
\usepackage{algorithmicx}%
\usepackage{algpseudocode}%
\usepackage{listings}%
\usepackage{subcaption}%
\usepackage{pifont}  
\usepackage{makecell}
\usepackage{tabularx} 
\usepackage{latexsym}
\usepackage{diagbox}
\usepackage{adjustbox}
\usepackage{caption}
\usepackage{soul}
\usepackage{stfloats}
\usepackage{threeparttable}
\usepackage{hyperref}
\usepackage{url}
\usepackage{array}
\usepackage{ragged2e}
\usepackage{enumitem}
\usepackage{microtype}

\newcommand{\Checkmark}{\ding{51}}   
\newcommand{\XSolidBrush}{\ding{55}} 

\makeatletter

\renewcommand{\makeheadbox}{}

\date{\mbox{}}

\patchcmd{\@maketitle}
  {\hrule\@height0.35mm\noindent}
  {\noindent}
  {}{}

\patchcmd{\@maketitle}
  {\vss\vskip19.5pt}
  {\vss\vskip0pt}
  {}{}

\patchcmd{\@maketitle}
  {\hrule\@height10.5mm\@width0\p@}
  {\hrule\@height0mm\@width0\p@}
  {}{}

\patchcmd{\@maketitle}
  {\noindent{\small\@date\if@twocolumn\vskip 7.2mm\else\vskip 5.2mm\fi}}
  {}
  {}{}

\makeatother

\begin{document}

\title{Towards Fine-Grained Text-to-3D Quality Assessment: A Benchmark and\\A Two-Stage Rank-Learning Metric}

\author{
Bingyang Cui$^{1,*}$ \and
Yujie Zhang$^{1,*}$ \and
Qi Yang$^{2,\dagger}$ \and
Zhu Li$^{2}$ \and
Yiling Xu$^{1,\dagger}$
}

\institute{
$^{1}$ Shanghai Jiao Tong University, China \\
\email{\{ccccby0813, yujie19981026, yl.xu\}@sjtu.edu.cn}\\
$^{2}$ University of Missouri-Kansas City, USA \\
\email{\{qiyang, lizhu\}@umkc.edu}\\
$^{*}$ Equal Contribution \\
$^{\dagger}$ Corresponding author
}

\maketitle

\begin{abstract}

Recent advances in Text-to-3D (T23D) generative models have enabled the synthesis of diverse, high-fidelity 3D assets from textual prompts. However, existing challenges restrict the development of reliable T23D quality assessment (T23DQA). 
First, existing benchmarks are outdated, fragmented, and coarse-grained, making fine-grained metric training infeasible. Moreover, current objective metrics exhibit inherent design limitations, resulting in non-representative feature extraction and diminished metric robustness. To address these limitations, we introduce \textbf{T23D-CompBench}, a comprehensive benchmark for compositional T23D generation. We define five components with twelve sub-components for compositional prompts, which are used to generate 3,600 textured meshes from ten state-of-the-art generative models. A large-scale subjective experiment is conducted to collect 129,600 reliable human ratings across different perspectives. Based on T23D-CompBench, we further propose \textbf{Rank2Score}, an effective evaluator with two-stage training for T23DQA. Rank2Score enhances pairwise training via supervised contrastive regression and curriculum learning in the first stage, and subsequently refines predictions using mean opinion scores to achieve closer alignment with human judgments in the second stage. Extensive experiments and downstream applications demonstrate that Rank2Score consistently outperforms existing metrics across multiple dimensions and can additionally serve as a reward function to optimize generative models. The project is available at \url{https://cbysjtu.github.io/Rank2Score/}.

\end{abstract}

\section{Introduction}\label{sec:introduction}

\begin{figure}[t]
  \centering
  \vspace{-0.8cm}
   \includegraphics[width=0.98\linewidth]{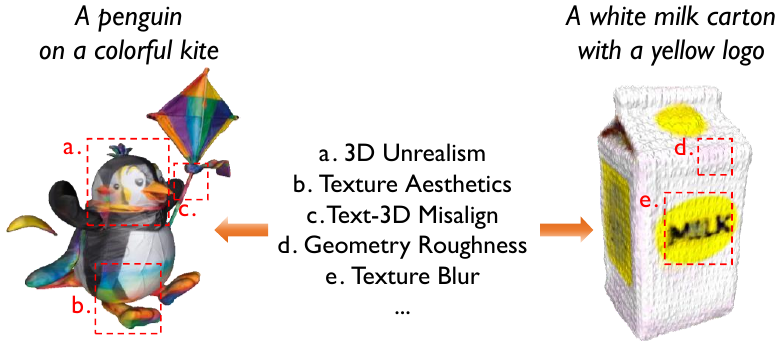}
   \setlength{\abovecaptionskip}{2pt} 
   \caption{Illustration of common distortions occurred in generated meshes.}
   \label{fig:dataset_motivation}
\end{figure}

Recently, Text-to-3D (T23D) generative models have made remarkable progress in producing diverse and high-fidelity 3D content from textual prompts \citep{zhao2025hunyuan3d, tripo, meshy, rodin, one2345++}. However, unlike traditional distortions in 3D data processing (\textit{e.g.}, noise or lossy compression \citep{graphsim}), the degradations in generated 3D assets are often unique, including Janus artifacts, semantic misalignment, and structural implausibility (see Fig. \ref{fig:dataset_motivation}), indicating current traditional quality assessment metrics cannot deal with generative distortion \citep{Geodesicpsim, msgeodesicpsim}. Therefore, T23D quality assessment (T23DQA) attracts considerable attention for ensuring the quality of experience in downstream applications \citep{xu2024imagereward, ye2024dreamreward, zhang2025refining, xu2024visionreward}. 

\begin{table*}[t]
\centering
\setlength{\abovecaptionskip}{1pt} 
\caption{Comparison of the existing T23D benchmarks. `\XSolidBrush' represents the scores are not available.}
\label{tab:dataset_comparison}
\resizebox{0.98\textwidth}{!}{
\begin{tabular}{c|c|c|c|c|c|c}
\toprule
Benchmark & \begin{tabular}[c]{@{}c@{}}Number of\\Prompt Categories\end{tabular} & \begin{tabular}[c]{@{}c@{}}Number of\\Rating Dimensions\end{tabular} & \begin{tabular}[c]{@{}c@{}}Number of\\Annotated Samples\end{tabular} & \begin{tabular}[c]{@{}c@{}}Number of\\Generative Models\end{tabular} & 3D Asset Type & Annotation Type\\ \midrule

\begin{tabular}[c]{@{}c@{}}$\mathrm{T}^3$Bench\\ \citep{he2023t3bench} \end{tabular} & 3 & 2 & 630 & 7 & Textured Mesh & \begin{tabular}[c]{@{}c@{}}\XSolidBrush\\(Absolute Score)\end{tabular}\\ \midrule

\begin{tabular}[c]{@{}c@{}}GPTEval3D\\ \citep{wu2024gpt4v} \end{tabular} & 2 & 5 & 234 pairs & 13 & 
\begin{tabular}[c]{@{}c@{}}NeRF, 3DGS,\\Textured Mesh\end{tabular}& \begin{tabular}[c]{@{}c@{}}\XSolidBrush\\(Preference Score)\end{tabular}\\ \midrule

\begin{tabular}[c]{@{}c@{}}MATE-3D\\ \citep{zhang2025hyperscore} \end{tabular} & 8 & 4 & 1,280 & 8 & Textured Mesh & \begin{tabular}[c]{@{}c@{}}\Checkmark\\(Absolute Score)\end{tabular}\\ \midrule

\begin{tabular}[c]{@{}c@{}}3DGCQA\\ \citep{zhou20253dgcqa} \end{tabular} & 10 & 2 & 313 & 7 & Textured Mesh & \begin{tabular}[c]{@{}c@{}}\Checkmark\\(Absolute Score)\end{tabular}\\ \midrule

\begin{tabular}[c]{@{}c@{}}AIGC-T23DAQA\\ \citep{fu2025multi} \end{tabular} & 23 & 3 & 969 & 7 & \begin{tabular}[c]{@{}c@{}}NeRF,\\Textured Mesh\end{tabular} & \begin{tabular}[c]{@{}c@{}}\Checkmark\\(Absolute Score)\end{tabular}\\ \midrule

\begin{tabular}[c]{@{}c@{}}GT23D-Bench\\ \citep{su2024gt23d} \end{tabular} & 7 & 10 & 504 & 6 & \begin{tabular}[c]{@{}c@{}}Point Cloud,\\Textured Mesh\end{tabular} & \begin{tabular}[c]{@{}c@{}}\XSolidBrush\\(Absolute Score)\end{tabular}\\ \midrule

\begin{tabular}[c]{@{}c@{}}T23D-CompBench\\ (proposed) \end{tabular} & 30 & 12 & 3,600 & 10 & Textured Mesh & \begin{tabular}[c]{@{}c@{}}\Checkmark\\(Absolute Score)\end{tabular}\\ \bottomrule
\end{tabular}}
\vspace{-0.5cm}
\end{table*}

Objective quality evaluation for T23D has been investigated for several years. Zero-shot metrics, including CLIPScore \citep{hessel2021clipscore}, BLIPScore \citep{li2022blip}, and pre-trained large language models (LLMs) \citep{he2023t3bench,wu2024gpt4v}, rely on image-text similarity or captioning. Some regression-based metrics employ hypernetworks \citep{zhang2025hyperscore} or multi-modal encoders \citep{fu2025multi} to extract representations and directly predict quality scores. Given the scarcity of available benchmarks, other works \citep{xu2024imagereward,ye2024dreamreward} employ ranking-based metrics, using pairwise comparisons to model relative preferences. Notably, the above metrics primarily focus on limited perspectives such as semantic alignment and overall quality. These coarse-grained perspectives can not quantify complex distortions in T23D generation such as texture blur, geometric loss, and surface roughness (see Fig. \ref{fig:dataset_motivation}) and also ignore other important quality factors like
authenticity and aesthetics. To provide a comprehensive understanding of generation quality, it is essential to develop a fine-grained evaluator that can detect specific quality perspectives of T23D generation.

\begin{figure}[t]
  \centering
   \includegraphics[width=0.98\linewidth]
   {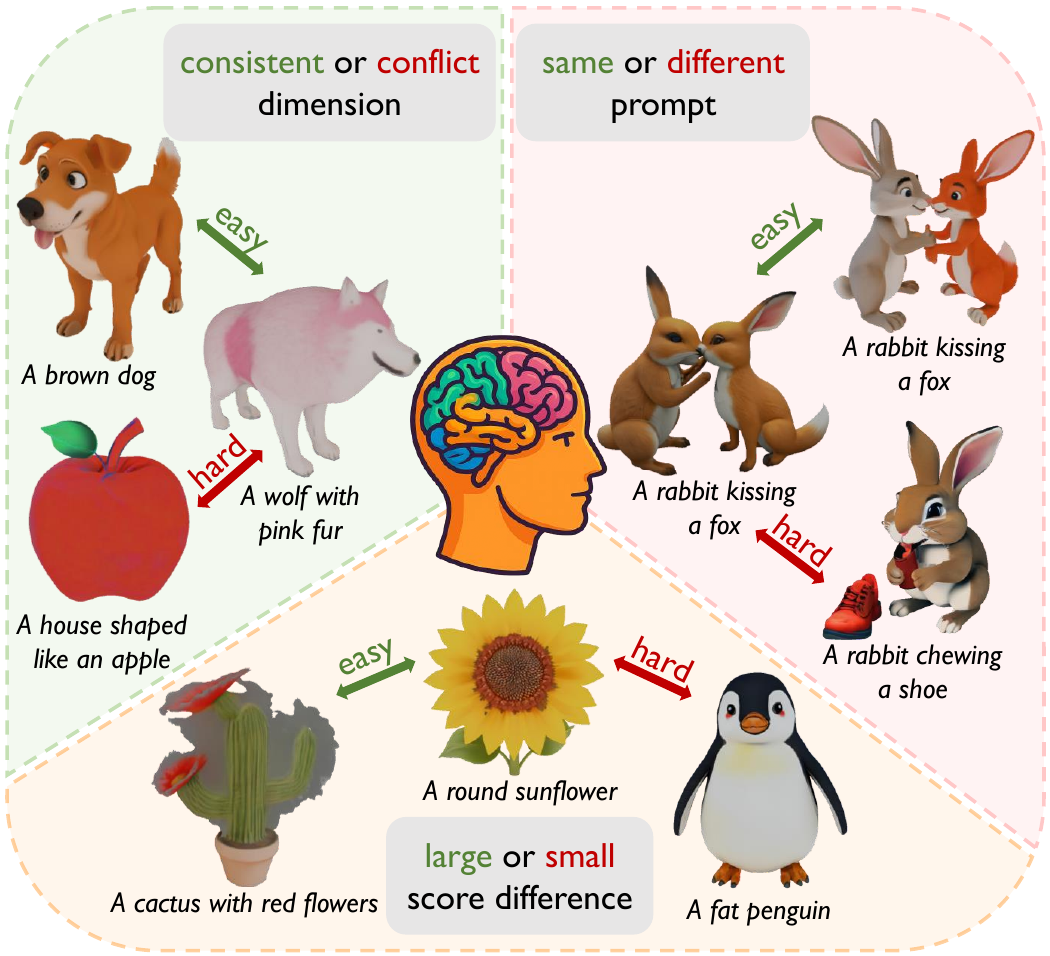}
   \setlength{\abovecaptionskip}{2pt} 
   \caption{Difficulty levels of pairwise ranking across prompts, scores, and dimensions.}
   \label{fig:metric_motivation}
\vspace{-0.6cm}
\end{figure}

However, there are still several challenges that restrict the development of fine-grained evaluators. First, current benchmarks are outdated, fragmented, and coarse-grained, limiting their utility for training reliable evaluators. As shown in Table \ref{tab:dataset_comparison}, these benchmarks rely on outdated generative models, and the prompt categories are often restricted to specific sub-problems \citep{wu2024gpt4v,he2023t3bench,zhou20253dgcqa,su2024gt23d}, overlooking compositional combinations among objects, attributes, relations, and styles \citep{zhang2025hyperscore,fu2025multi}. 
Moreover, they lack accurate annotations across different rating dimensions, making fine-grained evaluator training infeasible. Second, existing evaluators suffer from inherent design limitations. Zero-shot metrics \citep{liu2025clip, bai2025qwen2} often neglect visual or structural fidelity, while regression-based evaluators \citep{zhang2025hyperscore, fu2025multi} are constrained by limited benchmarks, leading to poorly unified feature representations. In comparison, ranking-based evaluators \citep{xu2024imagereward, ye2024dreamreward} can better exploit the quality differences among samples, but their performance typically decreases in more challenging scenarios. For example, they can easily give pairwise preferences to sample pairs with large quality differences, but usually fail when differences are small. As illustrated in Fig. \ref{fig:metric_motivation}, fine-grained T23DQA is influenced by multiple factors, including prompt identity, the differences of quality scores between samples, and the consistency of rankings across dimensions (\textit{e.g.}, sample A surpasses B in geometry but falls behind in texture). Ignoring these factors substantially reduces the robustness in assessing T23D quality.

To address the aforementioned limitations, we first introduce T23D-CompBench, a comprehensive benchmark designed for fine-grained evaluation of T23D generation. Considering the compositional complexity of natural language prompts, we define five components (\textit{i.e.}, \textit{Object}, \textit{Attribute}, \textit{Relationship}, \textit{Style}, and \textit{Length}) with 12 sub-components (as detailed in Fig. \ref{fig:combination_data_example}), each reflecting distinct semantic or structural aspects of 3D content. Based on these definitions, we use GPT-4o \citep{openaigpt} to generate $12\times30=360$ representative prompts that exhaustively cover all 30 component combinations. These prompts are fed into ten representative T23D generative models (\textit{e.g.}, Hunyuan-2.0 \citep{zhao2025hunyuan3d}, Tripo AI-2.5 \citep{tripo}, Meshy AI-5 \citep{meshy}), producing 3,600 3D assets. We then conduct a large-scale subjective study, where each mesh is annotated along twelve fine-grained dimensions (as detailed in Table \ref{tab:dimension_description}) by at least three human annotators, resulting in over $3,600 \times 12 \times 3 = 129,600$ reliable individual annotations. Finally, we conduct in-depth analyses of the collected scores to provide useful insights for future research in T23D generation.

Furthermore, we propose Rank2Score, a fine-grained ranking-based evaluator for T23DQA. We extract visual and textual features using a pre-trained CLIP model \citep{dosovitskiy2020image} and incorporate quality dimension-related texts through a projector. The fused cross-modal representation is then aligned with learnable prompts corresponding to different quality levels and dimensions, and the final quality score is derived based on their similarity. 
To enhance generalization and robustness, we adopt a \textit{two-stage} training strategy. \textbf{In the first stage}, we leverage pairwise ranking to learn relative preferences between samples. Considering that multiple factors significantly influence both the accuracy (see Table \ref{tab:curriculum_learning_motivation}) and the difficulty of ranking, we enhance pairwise training with curriculum learning \citep{wang2021curriculumsurvey}, where the evaluator is gradually exposed to increasingly difficult pairs. In addition, we incorporate a supervised contrastive regression loss \citep{zha2022supervised} to capture subtle distinctions between samples. This approach enables the evaluator to acquire robust and discriminative representations that effectively capture fine-grained variations across prompts and dimensions. 
\textbf{In the second stage}, the evaluator is fine-tuned directly on mean opinion score (MOS) labels to ensure closer alignment with human judgments. This process refines the evaluator for absolute quality estimation while preserving the relative ranking learned in the first stage. Experiments on four benchmarks demonstrate that Rank2Score outperforms existing metrics across all evaluation dimensions. 

Our contributions can be summarized as threefold:
\begin{itemize}
    \item We construct a large-scale benchmark T23D-CompBench, comprising 3,600 textured meshes generated by recent state-of-the-art models, along with 129,600 human ratings across twelve fine-grained quality dimensions. By incorporating diverse generative models, the benchmark covers a wider range of quality variations, offering a more representative basis for evaluation.
    \item We introduce Rank2Score, a two-stage evaluator that combines pairwise ranking with supervised contrastive regression and curriculum learning to achieve robust and fine-grained quality assessment. It jointly considers whether prompts are identical, the MOS difference between samples, and the coherence of rankings across dimensions, thereby yielding more accurate and reliable quality predictions.
    \item Extensive experiments demonstrate that Rank2Score outperforms existing metrics across multiple quality dimensions. An application of Rank2Score on the T23D model training further reveals that the proposed metric not only can predict accurate 3D content quality, but also can guide generative model training as supervision.
\end{itemize}

\section{Related Works}\label{sec:related work}

\subsection{T23D Generation}

T23D generative models aim to synthesize 3D content from textual prompts and have attracted increasing attention for their complexity and wide applicability. Refer to \citep{li2024advances,jiang2024survey,tang2025recent}, existing generation technologies primarily follow two research paradigms. The first paradigm, Optimization-based generation, focuses on integrating deep generative frameworks with diverse 3D representations to construct generalized 3D generation systems that support both unconditional and conditional generation. The second category, Feedforward generation, trains models using large-scale benchmarks and employs feedforward inference mechanisms to achieve the rapid and efficient generation of 3D assets. Notably, feedforward models substantially accelerate inference by eliminating the need for iterative optimization, thereby enabling the rapid generation of high-quality 3D content.

\begin{itemize}
\item \textbf{Optimization-based Generation.} These models typically leverage pre-trained multimodal networks to optimize 3D models based on user-specified prompts. DreamFusion \citep{dreamfusion} introduces this paradigm by optimizing Neural Radiance Fields (NeRF) using Score Distillation Sampling (SDS). TV-3DG \citep{yang2025tv} integrates visual prompt information to enhance generation quality. Follow-up models \citep{magic3d,latentnerf,wang2024prolificdreamer,seo2023let,ma2024scaledreamer,zhu2023hifa} adopt a two-stage coarse-to-fine strategy to improve both generation quality and efficiency. More recently, models such as \citep{ren2023dreamgaussian4d,ben2024dreamgaussian,yi2024gaussiandreamer,jaganathan2024ice,GaussianDreamerPro,di2025hyper} replace NeRF with 3D Gaussian Splatting (3DGS), achieving faster optimization with comparable fidelity. Additionally, several studies \citep{li2024instant3d,chen2024sculpt3d,yang2024viewfusion,seo2024retrievalaugmented,ye2023consistent1to3,liu2024vqa,luo2025trame} have explored fine-tuning pre-trained diffusion models to generate multi-view consistent images from a single input image. MVDream \citep{shi2024mvdream} constructs a diffusion model capable of generating multi-view images conditioned on text prompts. Zero-1-2-3 \citep{liu2023zero1to3} equips Stable Diffusion models with camera viewpoint control, enabling novel view synthesis from a single input image and specified camera transformations, which can be applied to 3D reconstruction tasks. One-2-3-45 \citep{liu2023one2345} and its derivative One-2-3-45++ \citep{one2345++} adopt Zero-1-2-3 as a multi-view diffusion model to generate 3D textured meshes from single images.
\item \textbf{Feedforward Generation.} Models such as Point-E \citep{nichol2022pointe} and Shap-E \citep{jun2023shape} directly learn from large-scale 3D benchmarks \citep{deitke2023objaversexl,deitke2023objaverse,su2024gt23d} to generate 3D assets in a feedforward manner. Despite their effectiveness, these models \citep{zhang2024clay,fu2022shapecrafter,gao2022get3d,zhang20233dshape2vecset,hessel2021clipscore,hong2023lrm,chen2025lara} are heavily reliant on 3D supervision, which remains limited in both scale and diversity. To address this challenge, LGM \citep{tang2024lgm} utilizes a latent Gaussian field, guided by 2D diffusion features, to enable controllable and scalable 3D synthesis in a purely feedforward manner. Hunyuan-2.0 \citep{zhao2025hunyuan3d} employs a dual-branch framework for shape and texture generation, enabling efficient feedforward reconstruction with strong semantic alignment. Similarly, Tripo AI-2.5 \citep{tripo} offers a commercial feedforward pipeline that enables text- or image-to-3D generation within seconds, supporting direct mesh export with PBR materials. Rodin Gen-1.5 \citep{rodin} builds upon triplane-based 3D-aware diffusion to achieve high-fidelity asset generation and editing with efficient inference. Meshy AI-5 \citep{meshy} provides a production-ready system, emphasizing rapid generation, view-consistent supervision, and prompt-based controllability, further advancing the practicality of feedforward paradigms in real-world creative workflows.

\end{itemize}

\subsection{T23D Benchmark}

Numerous benchmarks have been developed to assess the quality of T23D generative models. An overview of representative benchmarks is summarized in Table~\ref{tab:dataset_comparison}. These benchmarks differ across several aspects, including prompt design, evaluation dimensions, and benchmark scale.

For the prompt category, existing benchmarks focus on different aspects of prompt diversity. 3DGCQA \citep{zhou20253dgcqa} and GT23D-Bench \citep{su2024gt23d} target object category variation, while GPTEval3D \citep{wu2024gpt4v} emphasizes prompt creativity and complexity. $\mathrm{T}^3$Bench \citep{he2023t3bench} explores different object numbers and background contexts. MATE-3D \citep{zhang2025hyperscore} further proposes eight task-specific categories, including object attributes and spatial relationships, to capture compositional diversity. AIGC-T23DAQA \citep{fu2025multi} directly selects prompts from the PartiPrompt \citep{yu2022partiprompt} based on predefined categories and challenges, which are limited to individual aspects.

In terms of evaluation dimensions, \citep{zhou20253dgcqa,he2023t3bench,fu2025multi} primarily focus on alignment and overall quality. More recent studies, such as GPTEval3D \citep{wu2024gpt4v} and MATE-3D \citep{zhang2025hyperscore}, incorporate additional dimensions like geometry and texture, enabling more comprehensive assessments. However, these benchmarks evaluate these aspects at a holistic level and lack explicit measurements of fine-grained attributes (\textit{e.g.}, blur, roughness, authenticity, or aesthetics, see Fig. \ref{fig:dataset_motivation}). Although GT23D-Bench \citep{su2024gt23d} proposes ten rating dimensions, the benchmark is not publicly available and therefore cannot be utilized.

Regarding prompt categories and annotated samples, existing benchmarks employ early T23D models, such as DreamFusion \citep{dreamfusion}, Magic3D \citep{magic3d}, or Latent-NeRF \citep{latentnerf}, to generate objects. While these models play important roles in early development, their limited fidelity may constrain the utility of the benchmarks in supporting robust training or detailed quality analysis. The sizes of existing benchmarks also remain limited. Even the largest available benchmark, MATE-3D \citep{zhang2025hyperscore}, comprises only 1,280 annotated samples. Furthermore, benchmarks vary in output formats (\textit{e.g.}, meshes, NeRFs, point clouds), and some are not yet publicly released, limiting reproducibility and community adoption.

In summary, existing benchmarks provide valuable foundations for evaluating T23D generation, but there remains a need for a unified and high-quality benchmark that supports large-scale and fine-grained assessment with standardized data formats and high-fidelity assets.

\subsection{T23D Evaluator}

Evaluating T23D generative models is challenging as it requires both semantic understanding and accurate 3D perception. Early zero-shot metrics such as CLIPScore \citep{hessel2021clipscore} and BLIPScore \citep{li2022blip} rely on image-text feature similarity, but overlook visual or structural fidelity. Inspired by the rise of LLMs, \citep{wu2024gpt4v,he2023t3bench} leverage LLMs by combining them with vision encoders. However, LLMs are inherently optimized for reasoning rather than quality assessment, making direct application non-trivial and often requiring extensive retraining.

To more effectively address quality assessment, evaluators require fine-tuning on the corresponding benchmarks. Several regression-based evaluators have been proposed for multi-dimensional T23DQA. HyperScore \citep{zhang2025hyperscore} employs a hypernetwork to predict quality scores across multiple dimensions, while T23DAQA \citep{fu2025multi} adopts multimodal foundation models to jointly encode shape, texture, and text–3D consistency. Despite these advances, the limited scale of available benchmarks constrains regression-based training, which often yields poorly unified feature representations and thereby diminishes generalization performance on unseen data.

To mitigate data scarcity, ImageReward \citep{xu2024imagereward} and DreamReward \citep{ye2024dreamreward} adopt pairwise ranking to learn relative preferences. However, these evaluators often construct sample pairs from identical prompts, which may not fully capture the varying difficulty across different scenarios. As illustrated in Fig. \ref{fig:metric_motivation}, factors including prompt identity, the differences of quality scores between samples, and the consistency of rankings across dimensions all influence the difficulty of predicting pairwise preferences. Ignoring these relationships limits their robustness in assessing comprehensive quality.

In summary, existing evaluators are either narrowly focused on alignment or insufficiently fine-grained, underscoring the need for a more comprehensive and robust evaluator. To this end, we propose a two-stage framework for fine-grained quality assessment. Inspired by the curriculum learning \citep{wang2021curriculumsurvey} strategy, we organize the training process adaptively according to the pair difficulty, thereby enhancing the robustness and stability of the proposed evaluator.

\begin{figure*}[t]
  \centering
   \includegraphics[width=0.98\linewidth]{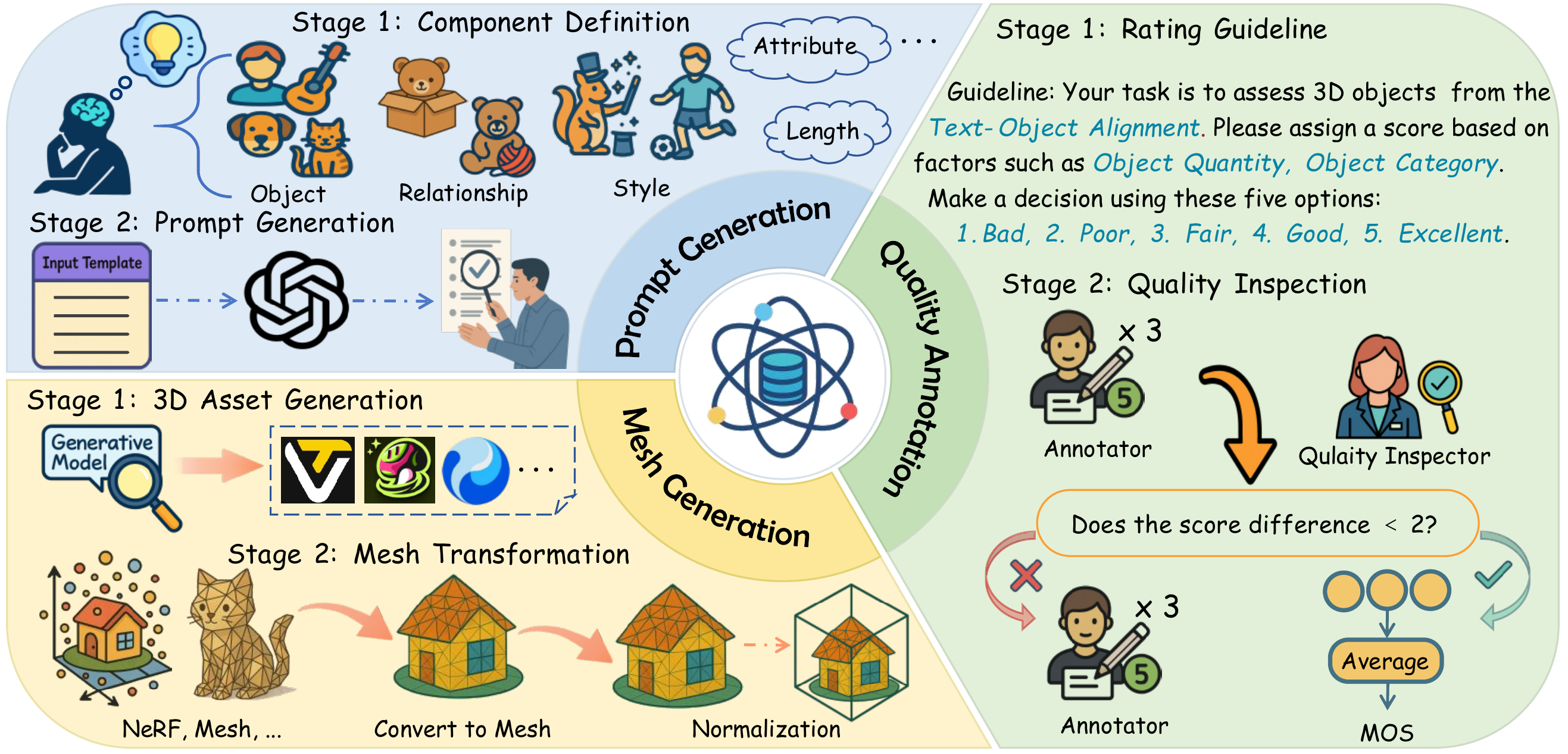}
   \setlength{\abovecaptionskip}{2pt}
   \caption{Illustration of the benchmark construction pipeline.}
   \label{fig:dataset_construction}
\vspace{-0.5cm}
\end{figure*}

\section{Benchmark Construction and Analysis}\label{sec:database}

In this section, we present the construction and analysis of our benchmark in detail. An overview of the benchmark construction pipeline is illustrated in Fig. \ref{fig:dataset_construction}. The benchmark consists of three main stages: prompt generation (Sec. \ref{sec:prompt_generation}), mesh generation (Sec. \ref{sec:mesh_generation}), and quality annotation (Sec. \ref{sec:quality_annotation}).  An extensive analysis of the collected MOS is presented in Sec. \ref{sec:observation} to demonstrate the rich visual characteristics and reliability of the proposed benchmark.

\subsection{Prompt Generation}\label{sec:prompt_generation}

To comprehensively evaluate the generative capabilities of T23D models, we design complex prompts with a diverse set of components encompassing various types and semantic concepts. Specifically, we introduce five components and twelve sub-components, each targeting a distinct aspect of compositional and semantic complexity in 3D generation, as detailed below:

(i) \textbf{Object.} This component focuses on the object categories and quantities described in the prompt. This component is divided into two sub-categories, \textit{Single} (\textit{e.g.}, \textit{``A black cat"}) and \textit{Multiple} (\textit{e.g.}, \textit{``An apple and a banana"}), based on the object quantity. (ii) \textbf{Relationship.} Each prompt contains at least two objects with specified relationships between the objects. Based on the type of relationships, we define two sub-categories: \textit{Spatial} (\textit{e.g.}, \textit{``A butterfly on a flower"}) and \textit{Non-spatial} (\textit{e.g.}, \textit{``A dog barking at a cat"}). (iii) \textbf{Style.} This component captures the stylistic intent of the generation, which can fall into two sub-categories: \textit{Realistic} (\textit{e.g.}, \textit{``A bronze statue"}), referring to plausible and real-world styles, and \textit{Imaginative} (\textit{e.g.}, \textit{``A cat with wings"}), referring to creative, fantastical, or physically implausible styles. (iv) \textbf{Attribute.} This component targets the specific descriptive attributes included in the prompt, such as geometry and appearance. It is divided into three sub-categories: \textit{Geometry-only} (\textit{e.g.}, \textit{``A curved brush"}), \textit{Appearance-only} (\textit{e.g.}, \textit{``A red coffee mug"}), and \textit{Mixed} (\textit{e.g.}, \textit{``A round teapot of rainbow colors"}). (v) \textbf{Length.} This component reflects the complexity and granularity of the prompt. We classify prompts into three sub-components based on their length and detail: \textit{Basic} (\textit{e.g.}, \textit{``A blue scarf"}), \textit{Refined} (\textit{e.g.}, \textit{``A red tiger with a wicker tail"}), and \textit{Complex} (\textit{e.g.}, \textit{``A spiky green iguana with a long tail and a row of spines along its back"}).

\begin{figure*}[t]
  \centering
   \includegraphics[width=0.96\linewidth]{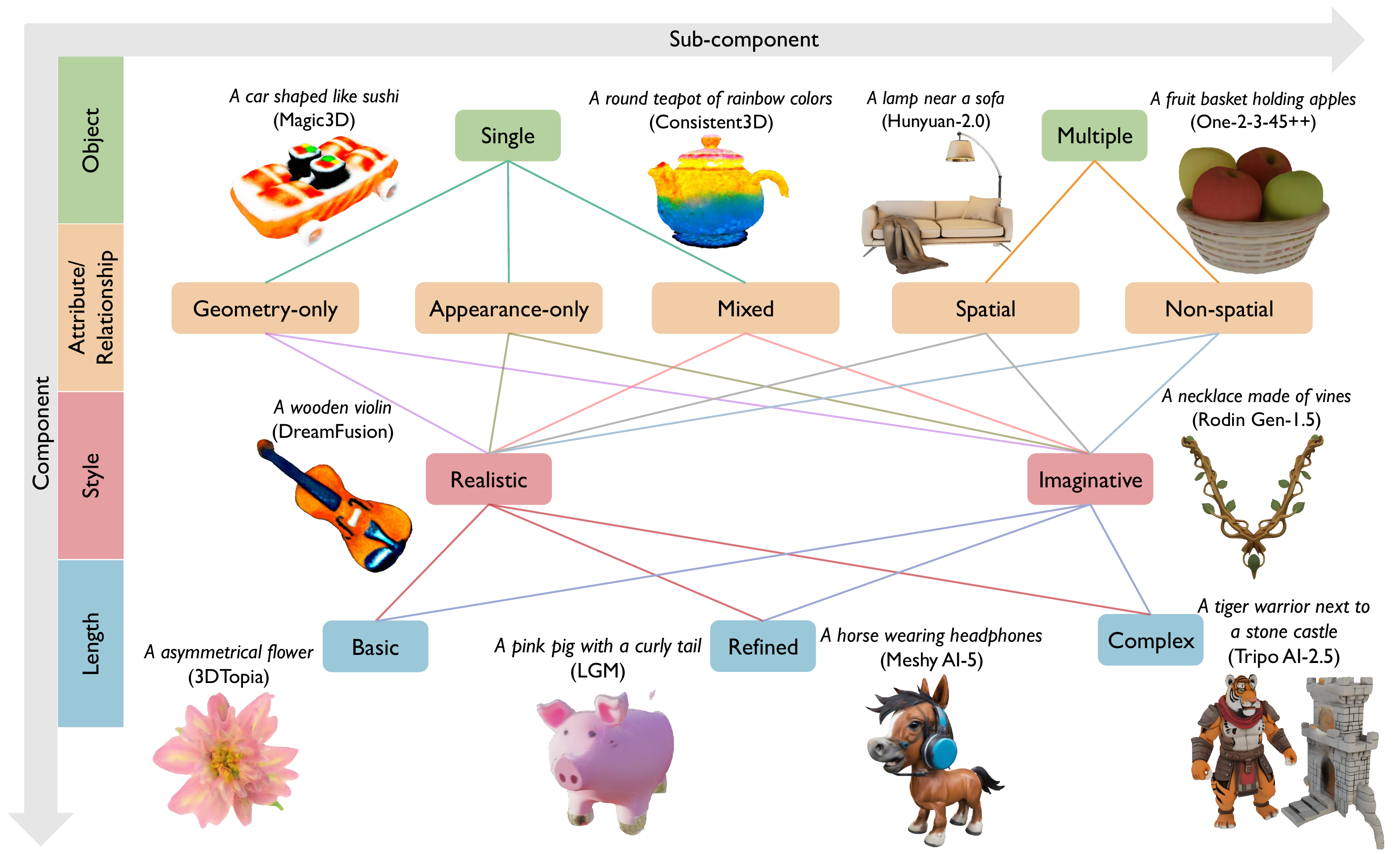}
   \setlength{\abovecaptionskip}{-2.5pt}
   \caption{Samples generated from different models and component combinations.}
   \label{fig:combination_data_example}
\vspace{-0.5cm}
\end{figure*}

Based on the defined components, we derive $(2+3)\times2\times3= 30$ possible combinations, where \textit{Single} focuses on \textit{Attribute} and \textit{Multiple} emphasizes \textit{Relationship} (see Fig. \ref{fig:combination_data_example}). For each combination, we employ GPT-4o \citep{openaigpt} to first generate 100 candidate prompts following \citep{wu2024gpt4v, zhang2025hyperscore}, and then filter out anomalous prompts (\textit{e.g.}, incoherent or incomprehensible ones, such as \textit{``A realistic cat made of water''}, \textit{``A large big huge massive elephant"}, \textit{``A wooden cloud with realistic texture"}, or \textit{``Blue red happy''}). The filtering process and representative examples are shown in Fig. \ref{fig:intruction_template2}. Ultimately, 12 valid prompts are retained for each combination, yielding a total of $12\times30 = 360$ prompts, which are evenly distributed across object categories. Detailed instructions are provided in Appendix \ref{app:instruction_of_prompt_pipeline}. Owing to the predefined pipeline, this process can be easily scaled to larger benchmarks or extended to additional combinations.

\subsection{Mesh Generation}\label{sec:mesh_generation}

After prompt generation, 3D assets are obtained using ten popular T23D models, which include both optimization-based models (\textit{e.g.}, DreamFusion \citep{dreamfusion}, Magic3D \citep{magic3d}, Consistent3D \citep{consistent3d}, One-2-3-45++ \citep{one2345++}) and feedforward models (\textit{e.g.}, 3DTopia \citep{3dtopia}, LGM \citep{tang2024lgm}, Tripo AI-2.5 \citep{tripo}, Rodin Gen-1.5 \citep{rodin}, Meshy AI-5 \citep{meshy}, and Hunyuan-2.0 \citep{zhao2025hunyuan3d}).

For each model, we utilize the official open-source implementation with default weights. The outputs are first converted into textured meshes and subsequently normalized, resulting in a total of $10 \times 360 = 3,600$ samples. Fig. \ref{fig:combination_data_example} illustrates representative meshes from the benchmark alongside their corresponding prompts and generative models.
\begin{table*}[t]
\centering
\setlength{\abovecaptionskip}{0pt} 
\caption{Descriptions for different sub-dimensions.}
\label{tab:dimension_description}
\resizebox{0.78\textwidth}{!}{
\begin{tabular}{c|c|c}
\toprule
\multirow{1}{*}{Dimension} & \multicolumn{1}{c|}{Sub-Dimension}  & \multicolumn{1}{c}{Description} \\ 
\midrule
\multirow{4}{*}{\begin{tabular}[c]{@{}c@{}}Textual-3D\\Alignment\end{tabular}} &Object Alignment (OA)	& category, quantity, count, type \\
& Attribute Alignment (AA) & material, geometry, appearance, shape \\
& Interaction Alignment (IA) & action, position, location, orientation \\
& Overall Alignment (OVA) & object, number, attribute, interaction \\ \midrule
 
\multirow{6}{*}{\begin{tabular}[c]{@{}c@{}}3D Visual\\Quality\end{tabular}} &Texture Clarity (TC)	& detail, resolution, visibility, contrast \\
 
&Texture Aesthetics (TA) & lighting, color, style, artistry \\
 
& Geometry Loss (GL) & incompleteness, integrity, fragmentation, infidelity \\
 
& Geometry Redundancy (GR) & overlap, floater, excess, duplication \\

& Geometry Roughness (GRS) & smoothness, irregularity, edge, crudeness \\

& Overall Visual (OV)
 & clarity, aesthetics, integrity, roughness \\ \midrule

\multirow{1}{*}{3D Authentic} & 3D Authentic (3DA)	& unrealism, inconsistency, overgeneration, implausibility
 \\ \midrule

 \multirow{1}{*}{Overall Quality} & Overall Quality (OQ)	& alignment, geometry, texture, authenticity
\\ \bottomrule
\end{tabular}}
\vspace{-0.3cm}
\end{table*}

\subsection{Quality Annotation}\label{sec:quality_annotation}

The quality of the generated textured mesh is influenced by multiple factors. To systematically capture human preferences, we propose a multi-dimensional rating framework consisting of four dimensions and twelve sub-dimensions, defined as follows:

(i) \textbf{Textual-3D Alignment.} This dimension evaluates the semantic consistency between the generated 3D meshes and the input prompt. We define 4 sub-dimensions that capture different aspects of alignment: \textit{Object Alignment (OA), Attribute Alignment (AA), Interaction Alignment (IA), and Overall Alignment (OVA)}. 
(ii) \textbf{3D Visual Quality.} To comprehensively assess the visual and structural fidelity of the generated 3D meshes, we introduce 6 sub-dimensions covering both texture and geometry: \textit{Texture Clarity (TC), Texture Aesthetics (TA), Geometry Loss (GL), Geometry Redundancy (GR), Geometry Roughness (GRS), and Overall Visual Quality (OV)}.
(iii) \textbf{3D Authentic (3DA).} This dimension assesses the realism and plausibility of the generated meshes, penalizing artifacts caused by symmetry or duplication (\textit{e.g.}, multiple faces or extra limbs) that are not justified by the prompt.
(iv) \textbf{Overall Quality (OQ).} Annotators are asked to provide an overall subjective rating by integrating their judgments across all the above dimensions. Table \ref{tab:dimension_description} summarizes the evaluation dimensions along with brief descriptions of their key focus. More comprehensive descriptions for each dimension are provided in Appendix \ref{app:definition_of_different_quality_dimensions}.

\begin{figure*}[t]
  \centering
   \includegraphics[width=\linewidth]{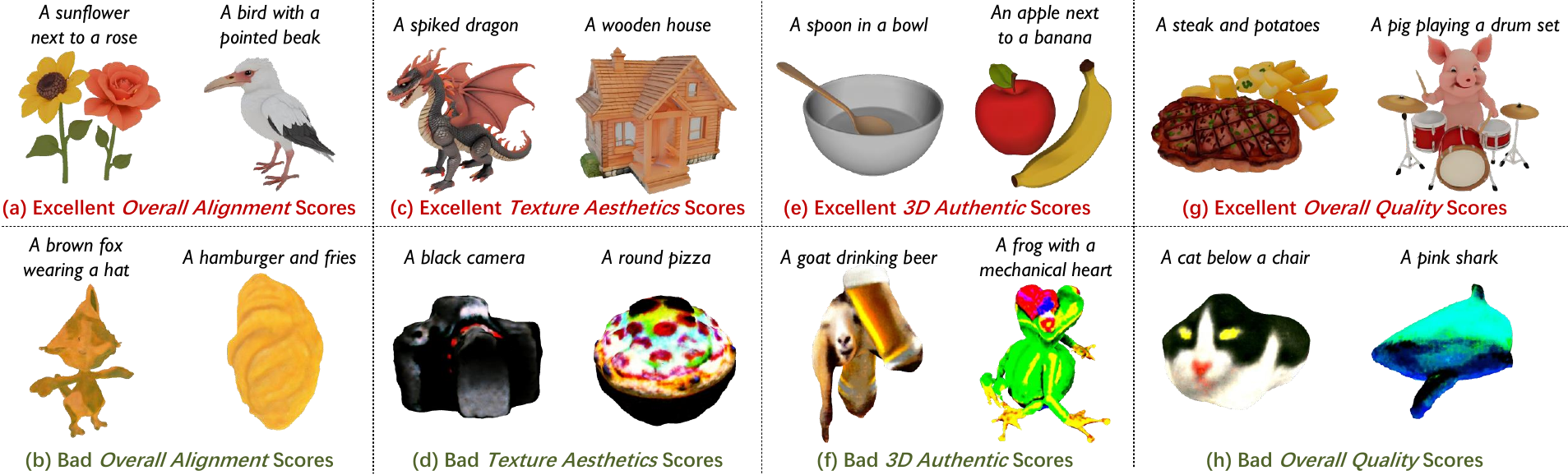}
   \setlength{\abovecaptionskip}{0pt}
   \caption{Different samples with excellent and bad in each quality dimension.}
   \label{fig:mos_example}
\vspace{-0.4cm}
\end{figure*}

To obtain the MOS of each mesh, we invite participants to evaluate the samples across the defined quality dimensions. We adopt the 5-point rating scale recommended by ITU-T P.910 \citep{p910} as the voting methodology. An interactive evaluation protocol is employed, allowing participants to freely adjust their viewing angles according to their preferences \citep{sjtupcqa,cui2024sjtu,yang2024benchmark}. Following \citep{wu2024gpt4v,huang2023t2i,sun2025t2v,han2024evalmuse}, each textured mesh is rated by three annotators across twelve evaluation dimensions. In cases where the scores from the three annotators exhibit significant disagreement (\textit{i.e.}, range $\ge $ 2), we conduct re-annotation until the scores fall within an acceptable agreement threshold \citep{TDMD, TSMD}. Ultimately, each sample is annotated by 3 annotators, and the average score is used as the MOS. Fig. \ref{fig:dataset_construction} illustrates the pipeline of quality annotation, while Fig. \ref{fig:mos_example} presents samples with extreme scores in selected dimensions. Details of the subjective evaluation environment and protocol are provided in Appendix \ref{app:subjective_experiment_procedure}.

\subsection{Observations}\label{sec:observation}

Based on the established benchmark, we conduct an in-depth analysis of the collected MOS from multiple perspectives as follows.

\subsubsection{Analysis of Quality Dimension} 
\begin{figure}[t]
  \centering
   \includegraphics[width=\linewidth]{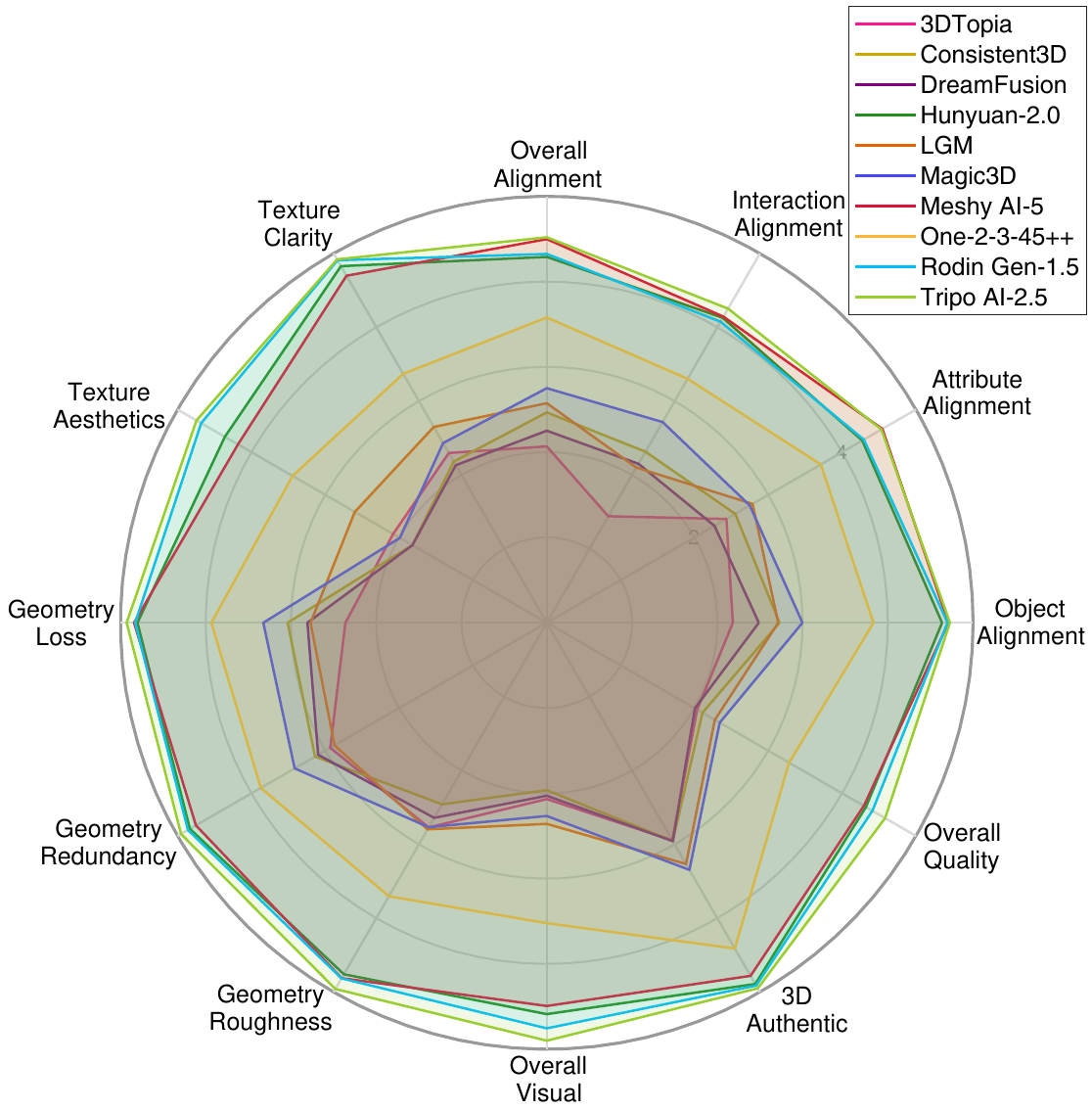}
   \setlength{\abovecaptionskip}{-1pt}
   \caption{MOS of different generative models across evaluation dimensions.}
   \label{fig:model_performance}
\vspace{-0.4cm}
\end{figure}

Fig. \ref{fig:model_performance} illustrates the average MOS over 3,600 samples of different generative models across evaluation dimensions. According to Fig. \ref{fig:model_performance}, the results demonstrate significant disparities across quality dimensions, with the worst dimension serving as the critical determinant of overall performance. Specifically, i) The score of \textbf{Alignment} category is the lowest among four dimensions, with \textit{IA} and \textit{AA} consistently underperforming across evaluated models. This indicates that current models struggle with complex prompts involving multiple objects and attributes. ii) For \textbf{Texture} quality, \textit{TA} are slightly lower than \textit{TC}, suggesting that while models generate clear textures, they still lack stylistic coherence. iii) For \textbf{Geometry} quality, higher scores in \textit{GL} and \textit{GR} indicate improvements in structural completeness, whereas \textit{GRS} remains a weaker aspect, reflecting challenges in surface smoothness. iv) The \textbf{Overall Visual} score reflects a bottleneck effect, often constrained by the weakest aspect of texture or geometry. v) The \textbf{Overall Quality} dimension is similarly affected by the weakest sub-dimension, acting as an aggregate reflection of failures across alignment, geometry, texture, and realism. This highlights the importance of consistent performance across dimensions for achieving high perceived quality.

\begin{table*}[t]
\centering
\setlength{\abovecaptionskip}{2pt}
\caption{Average scores of each generative model on twelve prompt components. The best model for each sub-component is marked in boldface, while the second is \underline{underlined}.}
\label{tab:componet_performance}
\resizebox{\textwidth}{!}{
\begin{tabular}{c|cc|cc|cc|ccc|ccc}
\toprule
\multirow{2}{*}{Model} & \multicolumn{2}{c|}{Object} & \multicolumn{2}{c|}{Relationship} &\multicolumn{2}{c|}{Style} & \multicolumn{3}{c|}{Attribute} & \multicolumn{3}{c}{Length} \\ \cmidrule(lr){2-13}
& Single & Multiple & Spatial & \makecell{Non-\\spatial}  & Realistic & Imaginative& \makecell{Geometry-\\only} & \makecell{Appearance-\\only} & Mixed & Basic & Refined & Complex \\ 
\midrule
3DTopia & 2.676 & 1.988 & 2.146 &1.837  &2.626  & 2.173 &2.730  &2.576  & 2.721 &2.389  & 2.431 & 2.383 \\
Consistent3D & 2.532 & 2.429 & 2.386 & 2.481 & 2.508 & 2.474 & 2.686 & 2.508 & 2.385 & 2.557 & 2.456 & 2.460 \\
DreamFusion  & 2.372 & 2.718 & 2.558 & 2.854 & 2.587 & 2.433 & 2.427 & 2.156 & 2.505 & 2.796 & 2.475 & 2.262 \\
Hunyuan-2.0      & 4.607 & 4.256 & 4.250 & 4.261 & 4.526 & 4.406 & 4.523 & 4.617 & 4.685 & 4.400 & 4.540 & 4.460 \\
LGM          & 2.880 & 2.464 & 2.448 & 2.489 & 2.716 & 2.711 & 2.677 & 2.981 & 2.975 & 2.728 & 2.843 & 2.572 \\
Magic3D      & 2.741 & 2.980 & 2.808 & 3.154 & 2.828 & 2.845 & 2.786 & 2.784 & 2.658 & 3.073 & 2.801 & 2.637 \\
Meshy AI-5        & 4.677 & \underline{4.337} & 4.324 & \underline{4.375} & 4.569 & 4.513 & 4.623 & 4.631 & 4.761 & \underline{4.521} & 4.637 & 4.467 \\
One-2-3-45++    & 3.746 & 3.732 & 3.643 & 3.835 & 3.820 & 3.660 & 3.748 & 3.566 & 3.932 & 3.868 & 3.786 & 3.569 \\ 
Rodin Gen-1.5       & \underline{4.725} & \textbf{4.590} & \textbf{4.473} & \textbf{4.719} & \textbf{4.712} & \textbf{4.629} & \underline{4.716} & \underline{4.680} & \underline{4.784} & \textbf{4.697} & \textbf{4.698} & \underline{4.617} \\
Tripo AI-2.5       & \textbf{4.809} & 4.333 & \underline{4.412} & 4.253 & \underline{4.684} & \underline{4.552} &\textbf{ 4.811 }& \textbf{4.752} & \textbf{4.862} & 4.475 & \underline{4.668} & \textbf{4.712} \\
\bottomrule
\end{tabular}}
\vspace{-0.4cm}
\end{table*}

\subsubsection{Analysis of Generative Models} 

From Fig. \ref{fig:model_performance}, we can see that optimization-based models generally show suboptimal performance, often placed in the innermost region of the radar chart. \textbf{DreamFusion} performs poorly across most dimensions, due to its reliance on NeRF, SDS, and Stable Diffusion for 2D supervision, which introduces instability and incoherent geometry. In contrast, \textbf{One-2-3-45++ } excels by integrating image-based supervision and multi-view diffusion models with controlled camera transformations, improving geometric fidelity and texture consistency. In comparison, feedforward models generally exhibit high generation performance. However, early models such as \textbf{3DTopia} and \textbf{LGM} were limited by the scale of 3D benchmarks used during training, resulting in poor performance. \textbf{Hunyuan-2.0} delivers strong and consistent performance across all dimensions. Its architecture comprises a shape module (Hunyuan3D-DiT), which employs Diffusion Transformers and Flow Matching for accurate mesh synthesis, and a texture module (Hunyuan3D-Paint) that combines geometry-aware, view-guided generation with heuristic viewpoint selection. Through dense view reasoning and texture baking, it achieves high texture fidelity and cross-view coherence. \textbf{Rodin Gen-1.5} utilizes triplane-based 3D-aware diffusion for high-fidelity generation and efficient editing. \textbf{Meshy AI-5} offers a production-ready system with rapid generation, view-consistent supervision, and prompt-based controllability, enhancing the real-world applicability of feedforward paradigms.

\subsubsection{Analysis of Prompt Component} 
We further report the average performance of each generative model across different prompt components in Table \ref{tab:componet_performance}. We have the following observations:
i) For the \textbf{Object} component, multi-object prompts pose greater challenges, primarily due to increased complexity in object interactions and occlusions.
ii) In the \textbf{Relationship} component, spatial and non-spatial prompts yield comparable results, suggesting that both types of relational understanding remain equally challenging for current models.
iii) For the \textbf{Style} component, models perform better on realistic prompts, likely due to training bias and limited generalization to imaginative or out-of-distribution concepts.
iv) Under the \textbf{Attribute} component, most models struggle more with appearance-only prompts, indicating relatively weaker texture modeling capabilities compared to geometry-oriented representations.
v) Finally, for the \textbf{Length} component, performance of most models tends to decline as prompt complexity increases, revealing difficulties in handling complex semantics and constraints. Notably, recent models like Tripo AI-2.5 maintain strong performance across all lengths, demonstrating robust generalization and instruction following.

Overall, models such as \textbf{Rodin Gen-1.5}, \textbf{Tripo AI-2.5}, \textbf{Meshy AI-5}, and \textbf{Hunyuan-2.0} demonstrate consistently strong and well-rounded performance across all prompt sub-components and evaluation dimensions. Their ability to integrate geometric accuracy, texture fidelity, and prompt understanding highlights their potential as strong reference models for future research in T23D generation.

\begin{figure*}[t]
  \centering
   \includegraphics[width=\linewidth]{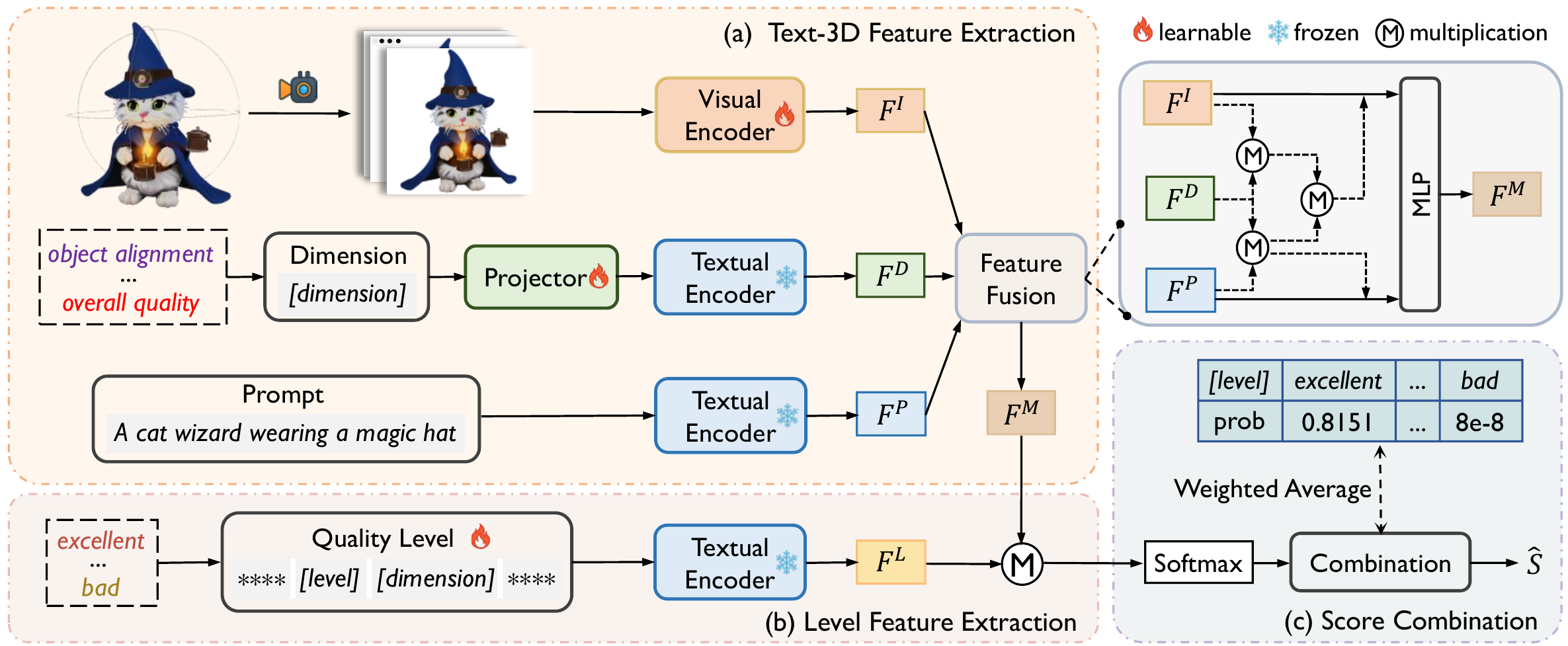}
   \setlength{\abovecaptionskip}{-3pt}
   \caption{The framework of the proposed Rank2Score.}
   \label{fig:framework}
\vspace{-0.5cm}
\end{figure*}

\section{Methodology}

Motivated by the aforementioned limitations of existing ranking-based evaluators, we introduce Rank2Score, a two-stage fine-grained evaluator. Our design is inspired by the principle of curriculum learning, which gradually increases the difficulty of samples during training. Specifically, Sec. \ref{sec:overall_framework} presents the overall framework of Rank2Score, while Sec. \ref{sec:training_procedure} describes the training procedure, including the two-stage training scheme and the three proposed curriculum learning strategies.

\subsection{Overall Framework}\label{sec:overall_framework}

The overall framework of our proposed Rank2Score is illustrated in Fig. \ref{fig:framework}, which comprises three main components: Text-3D feature extraction, level feature extraction, and score combination. Text-3D feature extraction module generates a unified cross-modal representation by integrating visual, textual, and dimensional information. Level feature extraction module learns representations for different quality levels. Finally, the score combination module measures the similarity between the cross-modal representation and the level features, yielding the final predicted quality score.

\subsubsection{Text-3D Feature Extraction}
Given a prompt $P$ and the corresponding mesh $\mathcal{M}$, we denote its multi-view texture maps as $I = \left \{I^{v} \in \mathbb{R}^{H\times W \times3}\mid_{v=1}^{N_v}   \right \} $, where $N_v$ represents the number of viewpoints. Let $D = \left \{ D^{d} \mid_{d=1}^{N_d}  \right \}$ represent the texts of multiple quality dimensions (\textit{e.g., ``object alignment''}), where $N_d$ represents the number of dimensions. We first employ visual and textual encoders from a transformer-based CLIP model \citep{dosovitskiy2020image}, denoted by $\mathcal{E}_{v}(\cdot )$ and $\mathcal{E}_{t}(\cdot )$, to obtain visual and textual features. To better capture local distortion patterns, we take the tokens from all image patches as the output of $\mathcal{E}_{v}(\cdot)$ and concatenate the visual features across $N_v$ viewpoints:
\begin{equation}
    F_{I}=\mathcal{E}_v(I)\in \mathbb{R}^{(N_v\times N_{t}^{v}) \times N_f}, F_{P}=\mathcal{E}_t(P)\in \mathbb{R}^{N_{t}^{p} \times N_f} ,
\end{equation}
where $N_{t}^{v}$ denotes the token number of the texture image (\textit{i.e.}, the patch number), $N_{t}^{p}$ denotes the token number of the prompt, and $N_f$ denotes the feature dimension.

To better distinguish different dimension features, we employ a projector module to map dimension texts (\textit{e.g.,} \textit{``object alignment"} and \textit{``overall quality"}) into richer representations. Then we feed the representations into the textual encoder to obtain dimension features:

\begin{equation}
F_{D}=\mathcal{E}_t(\mathcal{E}_p(D))\in \mathbb{R}^{N_d \times N_f} ,
\end{equation}
where $\mathcal{E}_p(\cdot)$ is a two-layer MLP serving as the projector.

Inspired by \citep{zhang2025hyperscore}, image patches from different viewpoints contribute unequally to the final prediction. Given the $d$-th dimension feature $F_{D}^d$, we compute the corresponding visual and textual weighting matrices as:
\begin{equation}
\begin{aligned}
&W_I^d=\text{SoftMax}((F_{I}\cdot (F_{P})^T)\cdot(F_{P}\cdot (F_{D}^d)^T)),\\
&W_P^d=\text{SoftMax}(F_{P}\cdot (F_{D}^d)^T),
\end{aligned}
\end{equation}
where $W_I^d$ captures the joint influence of each text token and the dimension token on every image patch, while $W_P^d$ encodes the relative importance of each text token with respect to the $d$-th dimension. Patches and tokens with stronger semantic relevance to the target dimension are assigned higher weights. Then the corresponding weighted visual and textual features for the $d$-th dimension are derived as:
\begin{equation}
\widetilde{F_I^d} = (W_I^d)^T \cdot F_{I} \in \mathbb{R}^{N_f}, \widetilde{F_P^d} = (W_P^d)^T \cdot F_{P} \in \mathbb{R}^{N_f}.
\end{equation}

Finally, we concatenate the weighted visual and textual features and pass them through a simple MLP to obtain the fused feature representation for the $d$-th dimension:
\begin{equation}
F_M^d = \text{MLP}(\widetilde{F_I^d}\oplus  \widetilde{F_P^d}) \in \mathbb{R}^{N_f},
\end{equation}
where $\oplus$ denotes the concatenation operation.

\subsubsection{Level Feature Extraction}
Considering that human beings prefer to describe quality using discrete quality descriptions rather than specific scores \citep{lyp-tbc, liu2025clip}, we adopt $N_l$ quality-related adjectives from ITU-R BT.500 \citep{bt500} (\textit{e.g.}, \textit{``excellent"}, \textit{``good"}, \textit{``fair"}, \textit{``poor"}, \textit{``bad"}) and combine them with $N_d$ dimension names to construct a set of quality level tokens and obtain their embeddings. For the $d$-th dimension, the quality level tokens are denoted as $L^d = \left \{ L_l^d \mid_{l=1}^{N_{l}}  \right \}$ (\textit{e.g.}, \textit{``good object alignment"}), where $l$ indexes the quality levels.

According to previous work \citep{liu2025clip}, prompts augmented with quality level tokens (\textit{e.g.}, ``this image has \textit{bad interaction alignment}" or ``an image of \textit{bad interaction alignment} quality") can be transformed into effective level features for quality assessment. To eliminate reliance on manually designed fixed prompts, we adopt a unified learnable prompt paradigm following CoOp \citep{zhou2022learning}, to automate prompt engineering. More specifically, quality level tokens $L$ are inserted into the middle of the template sentences, each of which consists of $K$ learnable context tokens, to construct the quality-related prompts $T_L$. These prompts $T_L$ are then fed into a frozen textual encoder to generate the corresponding level feature representations. For the $d$-th evaluation dimension, the level features are given by:
\begin{equation}
F_{L}^d=\mathcal{E}_t(T_L^d)\in \mathbb{R}^{N_l\times N_f}.
\end{equation}

\subsubsection{Score Combination}

After extracting the features, we compute the final quality score for each evaluation dimension. Specifically, we first calculate the cosine similarity between the fused feature $F_M^d$ and the level feature $F_L^d$, followed by softmax normalization:
\begin{equation}
\text{prob}^d=\text{SoftMax}\left (\frac{F_M^d\cdot (F_L^d)^T}{\left \| F_M \right \| \left \| F_L^d\right \|} \right )  \in \mathbb{R}^{N_l}.
\end{equation}
Each quality level is associated with a predefined quantitative value. For example, in benchmarks where MOS = 5 denotes the highest quality, the corresponding quality levels $\left[\textit{``excellent"}, \textit{``good"}, \textit{``fair"}, \textit{``poor"}, \textit{``bad"}\right]$ are mapped to $q = \left[5,4,3,2,1\right]$. Using the softmax probabilities as weights, the final predicted score for the $d$-th dimension is calculated by a weighted combination:
\begin{equation}
\hat{S^d }=\sum_{l=1}^{N_l} \text{prob}_l^d \cdot q _l.
\end{equation}

Finally, given $N_d$ evaluation dimensions, we obtain the predicted scores as $\hat{S}=\left \{ \hat{S^d } \mid_{d=1}^{N_d}  \right \}$, with the corresponding MOS denoted by $S=\left \{ S^d  \mid_{d=1}^{N_d}  \right \}$.

\subsection{Training Procedure}\label{sec:training_procedure}

\begin{figure}[t]
  \centering
   \includegraphics[width=0.96\linewidth]{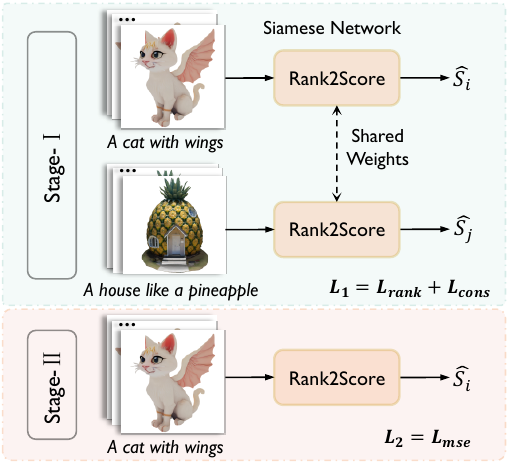}
   \setlength{\abovecaptionskip}{2pt}
   \caption{The training strategy of Rank2Score.}
   \label{fig:training_strategy}
\vspace{-0.5cm}
\end{figure}

To effectively optimize the network, we adopt a two-stage training strategy, as illustrated in Fig. \ref{fig:training_strategy}. \textbf{In the first stage}, we enlarge the training data by using a Siamese architecture \citep{chen2021exploring} to construct pairs and learn quality ranking. However, ignoring feature constraints at this stage may lead to insufficiently robust features. Consequently, we adopt a supervised contrastive regression loss \citep{zha2022supervised} that pulls features of pairs with similar MOS closer. Moreover, considering that the difficulty of rank comparison within a pair is influenced by multiple factors, we incorporate the principle of curriculum learning, progressively introducing training pairs from easy to hard to enhance robustness. \textbf{In the second stage}, the network is fine-tuned via score regression to better align the predicted score with the MOS.

\begin{figure}[t]
  \centering
   \includegraphics[width=\linewidth]{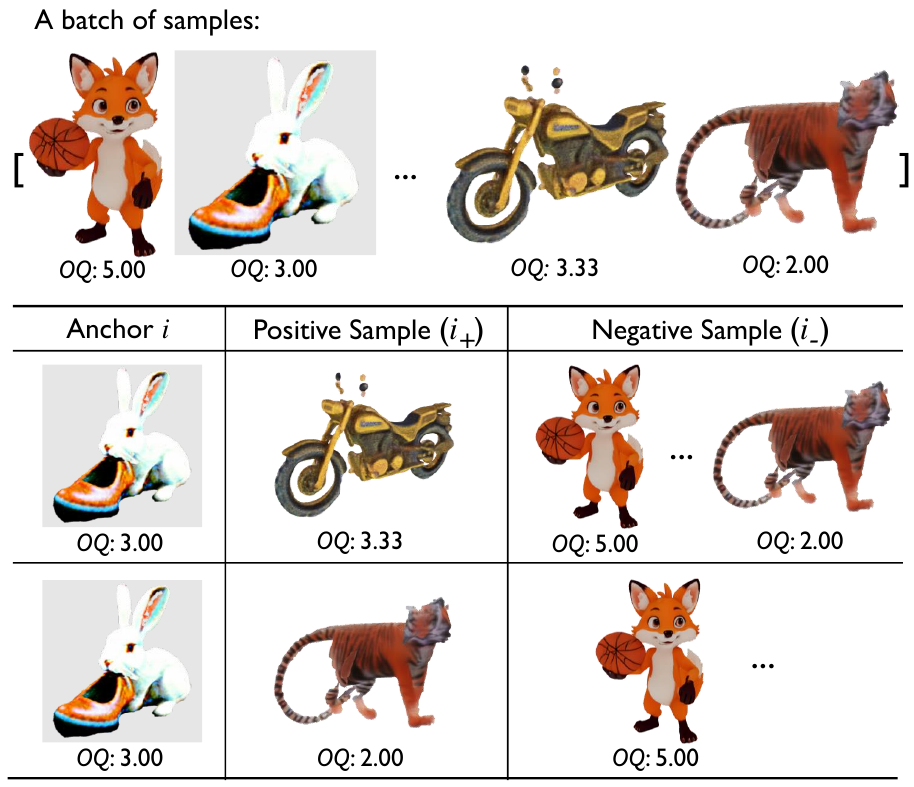}
   \setlength{\abovecaptionskip}{2pt}
   \caption{Illustration of $L_{cons}$. Example of positive and negative samples when the anchor is \textit{``rabbit"} (shown in gray shading). Taking \textit{OQ} as an example, when \textit{``motorcycle"} serves as the positive sample, their MOS label distance is 0.33, so the corresponding negative samples are \textit{``fox"} and \textit{``tiger"}, whose label distances to the anchor are 2.00 and 1.00 respectively. When \textit{``tiger"} represents the positive sample, their label distance is 1.00, and only \textit{``fox"}, which has a larger distance to the anchor, acts as the negative sample.}
   \label{fig:cons_loss}
\vspace{-0.5cm}
\end{figure}

\subsubsection{Stage-I: Ranking Learning based on Curriculum Learning}
To encourage the network to learn correct relative rankings, we employ a pairwise ranking hinge loss following \citep{liu2017rankiqa}. Given a mini-batch of $N_b$ samples, we construct all possible pairs and define the ranking loss as follows:
\begin{equation}
\begin{aligned}
&L_{rank} = \frac{2}{N_{b}(N_{b}-1)}
\sum_{\substack{i=1}}^{N_{b}}
\sum_{\substack{j \ne i}}
r(S_{i}, S_{j}, \hat S_{i}, \hat S_{j}) ,\\
&r(S_{i},S_{j}, \hat S_{i}, \hat S_{j})=\frac{1}{N_d}\sum_{\substack{d=1}}^{N_{d}}\max\limits \left( 0,\ -\eta_d \cdot(\hat S_{i}^d-\hat S_{j}^d) + \theta  \right).
\end{aligned}
\end{equation}
Here, $\eta_d = \text{sign}(S_{i}^d - S_{j}^d)$ indicates the ground-truth ranking direction for the $d$-th dimension, $i$ and $j$ index samples within the mini-batch that form a pair, and $\theta$ is a margin controlling the ranking tolerance.

Considering that pairs with large differences in MOS should be more distinguishable in the feature space \citep{shan2024contrastive}, we further introduce a supervised contrastive regression loss \citep{zha2022supervised}:
\begin{equation}
\begin{aligned}
&L_{cons} = \frac{1}{N_{b}(N_{b}-1)}\sum_{\substack{i=1}}^{N_{b}}\sum_{\substack{i_+ \ne i}}c(F_{M,i}, F_{M,i_+}, F_{M,i_-} ),\\
&c(i, i_+ , i_- )=-\frac{1}{N_{d}}\sum_{\substack{d=1}}^{N_{d}}\log_{}{\frac{\text{exp}(\text{sim}(i^d, i^d_+)/\tau )}{\sum_{\substack{i_{-} \ne i}} \delta \cdot \text{exp}(\text{sim}(i^d, i^d_- )/\tau)} },\\
&\delta  =
\left\{
\begin{array}{ll}
1, & \text{if } |S_i - S_{i_-}| \ge |S_i - S_{i_+}| \\
0, & \text{else} 
\end{array}\right.,
\end{aligned}
\end{equation}
where $F_{M,i}$ denotes the fused feature of the $i$-th sample, $i_+$ and $i_-$ represent the positive and negative samples corresponding to the $i$-th anchor, and $S_i$ denotes the MOS of the $i$-th sample. \text{sim}($\cdot$, $\cdot$) is the function measuring the similarity between two feature embeddings (\textit{i.e.},
negative L2 norm) and $\tau$ is the temperature parameter. The selection of positive and negative samples is illustrated in Fig. \ref{fig:cons_loss}. Through this procedure, the contrastive regression loss can pull closer the features of pairs with similar MOS while pushing apart those with larger MOS discrepancies, enabling the model to learn more robust representations. Accordingly, the overall loss function for the first stage is formulated as:
\begin{equation}
L_1= L_{rank} + \lambda L_{cons},
\end{equation}
where $\lambda$ is the weighting factor.

To address the varying difficulty levels of pairs and enhance ranking-based learning, we incorporate curriculum learning strategies that introduce training pairs from easy to hard, thereby facilitating smoother optimization and more robust model performance.

\begin{figure}[t]
  \centering
   \includegraphics[width=0.96\linewidth]{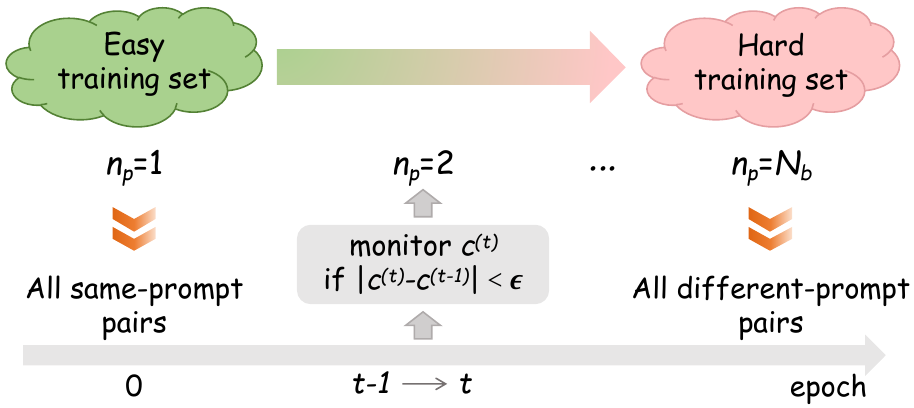}
   \setlength{\abovecaptionskip}{2pt}
   \caption{Strategy for Prompt Number.}
   \label{fig:strategy_prompt_number}
\vspace{-0.5cm}
\end{figure}

\textbf{Strategy for Prompt Number.} 
As mentioned in Fig. \ref{fig:metric_motivation}, pairs constructed from the same prompt are relatively easier to rank. We further validate the clarification in Table \ref{tab:curriculum_learning_motivation}, where all training pairs of the baseline are constructed under the same prompt. Comparing (1) and (2) in the table, we can see that the baseline provides inferior performance for sample pairs with different prompts, thus verifying our claim. To alleviate this problem, we adopt a curriculum learning strategy by gradually increasing the number of prompts per batch, thereby introducing training pairs from easy to hard. The strategy is illustrated in Fig. \ref{fig:strategy_prompt_number}. Specifically, we initialize with $n_p=1$, meaning that all pairs within a batch are generated from the same prompt. After each epoch, we monitor the model's performance on the training set. Let $\text{SRCC}^{d(t)}$ denote the Spearman Rank-Order Correlation Coefficient (SRCC) values for the $d$-th dimension at epoch $t$. The average SRCC across all $N_d$ dimensions is defined as:
\begin{equation}
    s^{(t)}=\frac{1}{N_d}  \sum_{d=1}^{N_d}\text{SRCC}^{d(t)}.
\end{equation}

If the performance improvement from epoch $t-1$ to epoch $t$ falls below a predefined threshold $\epsilon$, $n_p$ is incremented by 1 (up to the batch size $N_b$):
\begin{equation}\label{equ:strategy_1}
\left | s^{(t)}-s^{(t-1)} \right |  < \epsilon  \; \Rightarrow  \; n_p\gets \text{min}(n_p+1, N_b) .
\end{equation}
To ensure balanced sampling among $n_p$ prompts, the number of samples for the $p$-th prompt, denoted $n_s^p$, is defined as:
\begin{equation}
n_{s}^p =
\begin{cases}
\left\lfloor \dfrac{N_{b}}{n_p} \right\rfloor + 1, & \text{if } p \le N_{b} \bmod n_p \\\\
\left\lfloor \dfrac{N_{b}}{n_p} \right\rfloor, & \text{otherwise}
\end{cases}.
\end{equation}

\begin{figure}[t]
  \centering
   \includegraphics[width=0.94\linewidth]{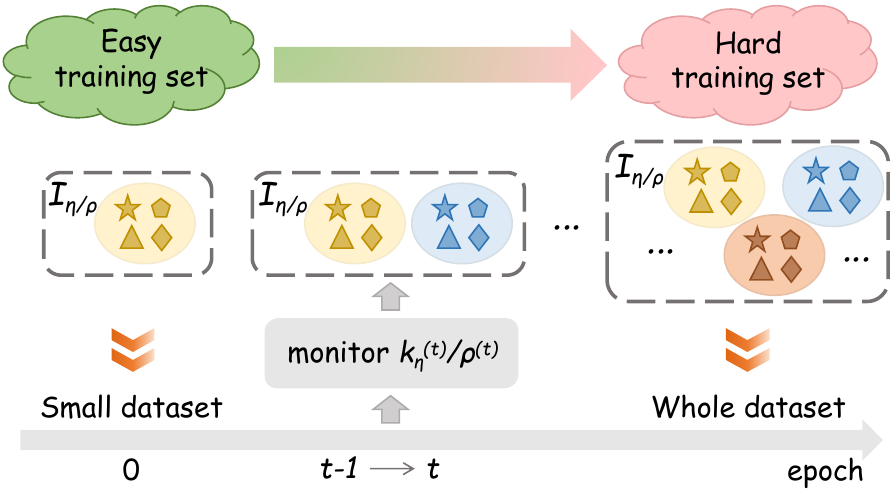}
   \setlength{\abovecaptionskip}{2pt}
   \caption{Strategy for Score Difference and Dimension Consistency.}
   \label{fig:strategy_score_difference}
\vspace{-0.5cm}
\end{figure}

\textbf{Strategy for Score Difference.} 
Seeing (3)-(6) in Table \ref{tab:curriculum_learning_motivation}, we can conclude that ranking difficulty decreases with increasing MOS differences. Building on this observation, we adopt a difference-aware curriculum strategy that progressively introduces harder pairs with smaller MOS differences, as illustrated in Fig. \ref{fig:strategy_score_difference}. Specifically, we define a difference threshold $\eta$ to control which pairs are used for training. For example, $\eta = 2.5$ means that only pairs with a score difference greater than 2.5 are included in the training process. Under this strategy, we evaluate ranking performance exclusively for pairs that satisfy $\eta$, rather than considering overall performance. Accordingly, we adopt Kendall’s Rank-Order Correlation Coefficient (KRCC), which directly quantifies the proportion of concordant and discordant pairs, to assess performance on the relatively small training subset. At training epoch $t$, the average KRCC is computed over the subset $\mathcal{I}_\eta$, which consists of all pairs satisfying the $\eta$ threshold:
\begin{equation}
k_\eta^{(t)} = \frac{1}{N_d} \sum_{d=1}^{N_d} \text{KRCC}(\mathcal{I}_\eta)^{d(t)},
\end{equation}
where $\text{KRCC}^{d(t)}$ denotes the KRCC for the $d$-th dimension at epoch $t$. Similar to $s^{(t)}$, if the performance improvement of $k_\eta^{(t)}$ between two consecutive epochs falls below $\epsilon$, $\eta$ is decreased by 1 to introduce more training pairs that meet the difference requirement as follows:
\begin{equation}\label{equ:strategy_2}
\left | k_\eta^{(t)}-k_\eta^{(t-1)} \right |  < \epsilon  \; \Rightarrow  \; \eta \gets \text{max}(\eta-1, 0).
\end{equation}
This process continues until all pairs are included in training. To mitigate the risk of data sparsity during the early stages of training, $\eta$ is initialized to half of the maximum MOS value.

\textbf{Strategy for Dimension Consistency.} As mentioned above, a pair becomes easier to rank when one sample consistently outperforms the other across all dimensions. To mitigate the interference arising from multi-dimensional training, we adopt a dimension-aware curriculum learning strategy. Specifically, for a pair of samples $(i, j)$ with MOS values ${S_i^d}$ and ${S_j^d}$ across $N_d$ dimensions, the consistency $c(i,j)$ is defined as:
\begin{equation}
\begin{aligned}
&c(i,j)=\frac{ \left |{\sum_{d=1}^{N_d}} \phi ^d(i,j)  \right |} {{\sum_{d=1}^{N_d}} \left | \phi ^d(i,j) \right | } ,\\
&\phi ^d(i,j)=\left\{\begin{matrix}
 +1, & S_i^d>S_j^d\\
 -1, & S_i^d<S_j^d\\
  0,& S_i^d=S_j^d
\end{matrix}\right..
\end{aligned}
\end{equation}
Here,$c(i,j)$ ranges from 0 to 1. $c(i,j)=1$ indicates complete consistency across all dimensions, making the pair easy to rank, whereas $c(i,j)=0$ corresponds to balanced comparison signs across dimensions, indicating maximal ranking difficulty. Given that dimension consistency has a relatively moderate impact (see Table \ref{tab:curriculum_learning_motivation} (7) and (8)), we adopt a simple linear schedule to progressively expand the training data, balancing computational cost and model performance. The process is illustrated in Fig. \ref{fig:strategy_score_difference}. Specifically, we define a dynamic threshold $\rho^{(t)}$:
\begin{equation}\label{equ:strategy_3}
\rho ^{(t)} = 0.5 - 0.5 \cdot \frac{t}{T}.
\end{equation}
Only pairs satisfying $c(i,j) \ge \rho^{(t)}$ are included in $\mathcal {I}_\rho$, allowing the model to initially focus on highly consistent pairs and gradually incorporate more diverse ones as training progresses.

\subsubsection{Stage-II: Regressive Learning}
Considering that the first stage only focuses on relative ranking between pairs, in the second stage, we further fine-tune the network using a regression loss based on the mean squared error (MSE) to better align the predicted scores with the MOS:
\begin{equation}
    L_{2} = L_{mse} = \frac{1}{N_bN_{d}}
    \sum_{i=1}^{N_b}
    \sum_{\substack{d=1}}^{N_{d}} (S_{i}^d - \hat S_{i}^d)^2.
\end{equation}

\section{Experiments}
\begin{table*}[t]
\centering
\caption{Performance Comparison on the proposed T23D-CompBench. The best metric in each column is marked in \textbf{boldface} while the second is \underline{underlined} (AP denotes average performance across all quality dimensions).}
\setlength{\abovecaptionskip}{-2pt}
\label{tab:overall_performance1}
\resizebox{\textwidth}{!}{
\begin{tabular}{c|c|c|c|c|c|c|c|c|c|c|c|c|c|c}
\toprule
Type & Metric & AP & OA & AA & IA & OVA & TC & TA & GL & GR & GRS & OV & 3DA  & OQ  \\ \midrule

& CLIPScore \citep{hessel2021clipscore} & 0.532 & 0.555 & 0.504 & 0.587 	& 0.561	& 0.548	& 0.523	& 0.520	& 0.475	& 0.509	& 0.547	& 0.492	&0.562
\\
&BLIPScore \citep{li2022blip} & 0.581 & 0.620 & 0.564 & 0.667	& 0.627	& 0.587	& 0.573	& 0.529	& 0.508	& 0.555	& 0.590	& 0.545	& 0.610
\\
&Aesthetic Score \citep{schuhmann2022laion} & 0.480 &0.455	& 0.385	& 0.400	& 0.423	& 0.542	& 0.530	& 0.493	& 0.492	& 0.514	& 0.536	& 0.480	& 0.510
\\
&ImageReward \citep{xu2024imagereward} & 0.678 & 0.724 & 0.651 & \underline{0.748} & 0.728	& 0.657	& 0.651	& 0.601	& 0.570	& 0.622	& 0.671	& 0.622	& 0.894
\\
&DreamReward \citep{ye2024dreamreward} & 0.638 & 0.692	& 0.604	& 0.685	& 0.678	& 0.647	& 0.641	& 0.594	& 0.561	& 0.613	& 0.659	& 0.605	& 0.682
\\
Zero-shot &HPS v2 \citep{wu2023human} & 0.688	&0.714  & 0.659	& 0.681	& 0.709	& 0.716	& 0.689	& 0.671	& 0.642	& 0.682	& 0.717	& 0.650	& 0.727
\\
&Q-Align \citep{wang2023exploring} & 0.709 & 0.667	& 0.591	& 0.543	& 0.626	& 0.822	& 0.813	& 0.700	& 0.702	& 0.791	& 0.806	& 0.689	& 0.753
\\
& LLaVA-NeXT (8B) \citep{li2024llava} & 0.361	& 0.475	& 0.441&0.512 & 0.400	& 0.498	& 0.219	& 0.209	& 0.215	& 0.217	& 0.230	& 0.451	& 0.460
\\ 
& mPLUG-Owl3 (7B) \citep{ye2024mplug} & 0.401 & 0.573	&0.540 	&0.579 &0.428	&0.579 	&0.251 	&0.210 	&0.211 	& 0.222	& 0.239	&0.486 	&0.490 
\\ 
& Qwen2.5-VL (7B) \citep{bai2025qwen2}  &0.575 &0.616	& 0.567	& 0.617	& 0.603	& 0.555	& 0.584	& 0.547	& 0.500	& 0.552	& 0.586	& 0.547	& 0.620
\\
& InternVL2.5 (8B) \citep{chen2024expanding} & 0.452	& 0.506&	0.502& 0.559	&0.492 	&0.450 	&0.431 	&0.392 	&0.407 	&0.401 	& 0.419	& 0.414	&0.451 
\\
& DeepSeekVL (7B) \citep{lu2024deepseek} & 0.565	& 0.585	&0.636 &	0.546&0.573 	& 0.545	&0.561 	&0.537 	&0.558 	&0.552 	&0.565 	&0.562 	&0.554 
\\ \midrule
& ResNet50 \citep{he2016deep} & 0.738	& 0.749 &	0.470 & 0.479	&0.694 	& 0.856	& 0.855	& 0.770	& 0.771	& 0.821	& 0.856	& 0.743	& 0.793
\\
& ViT-B \citep{dosovitskiy2020image} & 0.740	& 0.728	& 0.519 & 0.486	& 0.671	& 0.862	& 0.859	& 0.771	& 0.751	& 0.825	& 0.870	& 0.722	& 0.817 
\\
& SwinT-B \citep{liu2021swin} & 0.755	& 0.702 	& 0.543	& 0.463	& 0.693 & 0.878 	& 0.881	& 0.801	& 0.784	& 0.844	& 0.883	& 0.749	& 0.837
\\
Fine-tune & DINO v2 \citep{oquab2023dinov2} &0.704 &0.771	&0.576 	&0.573 	&0.713 	&0.887 	& 0.888	&0.816 	&0.712 	&0.849 	&0.897 	&0.764 	&0.845 
\\
&ImageReward \citep{xu2024imagereward} & 0.777	& 0.788	& \underline{0.675} & 0.727	& 0.748	& 0.830	& 0.812	& 0.773	& 0.756	& 0.792	& 0.837	& 0.748	& 0.832
\\
&HyperScore \citep{zhang2025hyperscore} & \underline{0.809}	& \underline{0.821} & 0.669	& 0.628	&\underline{0.794} 	&\underline{0.889} 	&\underline{0.889} 	&\underline{0.822} 	&\underline{0.800} 	&\underline{0.863} 	&\underline{0.906} 	&\underline{0.772} 	&\underline{0.860} 
\\ \midrule
proposed   & \textbf{Rank2Score} &\textbf{0.908 }&\textbf{0.912} & \textbf{0.880} 	&\textbf{0.907} 	&\textbf{0.928} 	& \textbf{0.920} &\textbf{0.944} 	&\textbf{0.898} &\textbf{0.878}  	&\textbf{0.887} &\textbf{0.948} &\textbf{0.862}  &\textbf{0.934}
\\ \bottomrule
\end{tabular}}
\vspace{-0.4cm}
\end{table*}
\subsection{Benchmarks and Evaluation Metrics}

To illustrate the effectiveness of our metric, we evaluate it on four benchmarks with available raw opinion scores: 3DGCQA \citep{zhou20253dgcqa}, MATE-3D \citep{zhang2025hyperscore}, T23DAQA \citep{fu2025multi}, and the proposed T23D-CompBench. The basic statistics of these benchmarks are summarized in Table \ref{tab:dataset_comparison}.

To better compare the overall ranking performance, we adopt SRCC as the metric, where higher values indicate better consistency with human judgments. Following \citep{fittingfunction}, the nonlinear logistic-5 function is applied to map the predicted scores to the MOS range, compensating for potential scale misalignment.

\subsection{Implement Details}\label{sec:implement_details}

\subsubsection{Benchmark Split} 
We apply a 5-fold cross-validation for all benchmarks while ensuring that there is no prompt overlap between the training and testing sets. For each fold, the performance on the test set with minimal training loss is recorded, and the average performance across all folds is recorded as the final result.

\subsubsection{Network Details} 
We use the Vision Transformer \citep{dosovitskiy2020image} with 16×16 patch embeddings (ViT-B/16) as the visual encoder in CLIP-Visual, and the pre-trained transformer in CLIP-Text as the textual encoder. All feature representations (visual, textual, dimensional, and level features) have a dimension of $N_f=512$. All textured meshes are rendered into $M=6$ projected images with a spatial resolution of 512 × 512 by PyTorch3D and then resized into a resolution of $224\times224$ as inputs \citep{yang2022no}. $\theta$ is set to 0.5, $\tau$ is set to 2, $\lambda$ is set to 1, and $\epsilon $ is set to $1e-2$.

\subsubsection{Training Strategy} 
We implement our metric using PyTorch and perform all experiments on NVIDIA RTX 3090 GPUs. The training is conducted in two stages. In the first stage, the textual encoder is frozen, and the network is trained for 40 epochs with a batch size of 8. In the second stage, both the textual encoder and the quality-level learner are frozen, and the network is trained for 10 epochs with the same batch size. We use the Adam optimizer \citep{kingma2014adam} with a weight decay of $1e-4$. The learning rate is set to $2e-6$ for the pre-trained visual encoder and $3e-4$ for all other parts, decaying by a factor of 0.9 every 5 epochs.

\subsection{Performance Comparison}

\begin{table*}[t]
\centering
\caption{Performance Comparison on the other three benchmarks. The best metric in each column is marked in \textbf{boldface} while the second is \underline{underlined}.}
\setlength{\abovecaptionskip}{0pt}
\label{tab:overall_performance2}
\resizebox{\textwidth}{!}{
\begin{tabular}{c|c|c|c|c|c|c|c|c|c|c}
\toprule
\multicolumn{2}{c|}{Benchmark} & \multicolumn{2}{c|}{3DGCQA} & \multicolumn{3}{c|}{AIGC-T23DAQA} & \multicolumn{4}{c}{MATE-3D} \\ \midrule
Type & Metric & OVA & OQ & OVA & 3DA & OQ & OVA & T & G & OQ  \\ \midrule
& CLIPScore \citep{hessel2021clipscore} 	& 0.356	& 0.360	& 0.631	& 0.500	& 0.592	& 0.494	& 0.537	& 0.496	& 0.510
\\
&BLIPScore \citep{li2022blip} 	& 0.376	& 0.398	& 0.612	& 0.454	& 0.552	& 0.533	& 0.578	& 0.542 & 0.554
\\
&Aesthetic Score \citep{schuhmann2022laion}  & 0.053	& 0.084	& 0.341	& 0.209	& 0.382	& 0.099	& 0.172	& 0.160	& 0.150
\\
&ImageReward \citep{xu2024imagereward}  	& 0.426	& 0.421	& 0.660	& 0.512	& 0.659	& 0.651	& 0.591	& 0.612	& 0.623
\\
&DreamReward \citep{ye2024dreamreward}  & 0.294 & 0.281 & 0.663	& 0.503	& 0.650	& 0.526	& 0.513	& 0.501	& 0.524
\\
Zero-shot &HPS v2 \citep{wu2023human} & 0.312	& 0.291	& 	0.638 & 0.471 & 0.680	& 0.416	& 0.423	& 0.412	& 0.420
\\
&Q-Align \citep{wang2023exploring} & 0.063	& 0.003		& 0.144	& 0.234	& 0.391	& 0.199	& 0.355	& 0.431	& 0.340
\\
& LLaVA-NeXT (8B) \citep{li2024llava} &0.215 	&0.197 	& 0.526	& 0.446	& 0.598	& 0.329	& 0.272	& 0.328	& 0.267
\\ 
& mPLUG-Owl3 (7B) \citep{ye2024mplug} & 0.416 	& 0.373  & 0.516	& 0.444	& 0.541	& 0.426	& 0.381	& 0.451	& 0.461
\\ 
& Qwen2.5-VL (7B) \citep{bai2025qwen2}  &0.495 	& 0.482	& 0.637	& 0.496	& 0.372	& 0.568	& 0.251	& 0.558	& 0.487
\\
& InternVL2.5 (8B) \citep{chen2024expanding} & 0.374	&0.312 	& 0.583	& 0.551	& 0.525	& 0.501	& 0.345	&0.421 	& 0.279
\\
& DeepSeekVL (7B) \citep{lu2024deepseek} & 0.401	& 0.494	& 0.587	& 0.586	& 0.558	& 0.597	& 0.543	& 0.502	& 0.431
\\ \midrule
& ResNet50 \citep{he2016deep} & 0.355 & 0.392 & 0.647	& 0.619	& 0.737	& 0.551	& 0.672	& 0.653	& 0.632
\\
& ViT-B \citep{dosovitskiy2020image} & 0.322	& 0.376	& 0.572	& 0.624	& 0.715	& 0.515	& 0.676	& 0.642	& 0.614
\\
& SwinT-B \citep{liu2021swin} & 0.466 & 0.480 & 0.609	& 0.640	& 0.705	& 0.506	& 0.640	& 0.610	& 0.592
\\
Fine-tune & DINO v2 \citep{oquab2023dinov2} & 0.421	& 0.486	& 0.685	& 	0.666 & 0.789	& 0.642	& 0.771	& 0.739	& 0.728 
\\
&ImageReward \citep{xu2024imagereward} & \underline{0.501}	& \underline{0.522}	& 0.665	&0.652 &	0.785& 0.658	& 0.686	&0.673 	& 	0.679
\\
&HyperScore \citep{zhang2025hyperscore} & 0.402	& 0.452	& \underline{0.746}	& \underline{0.674}	& \underline{0.810}	& \underline{0.739}	&\underline{0.811}	& \underline{0.782}	& \underline{0.792}
\\ \midrule
proposed   & \textbf{Rank2Score}	& \textbf{0.687}  & \textbf{0.727} 	& \textbf{0.872}	& \textbf{0.761}	& \textbf{0.840} & \textbf{0.868} &\textbf{0.894} & \textbf{0.898}  & \textbf{0.897}
\\ \bottomrule
\end{tabular}}
\vspace{-0.4cm}
\end{table*}

\subsubsection{Quantitative Results}

To comprehensively evaluate the performance of the proposed evaluator, we compare it with several state-of-the-art baselines, including twelve zero-shot (covering both pretrained image quality evaluators and LLM-based evaluators) and six fine-tuned quality assessment metrics. For fine-tuned metrics, we equip these prevalent backbones with multiple regression heads and fine-tune them using the same training protocol, including identical data splits and training settings (e.g., six-view projections, 224×224 input resolution, and a batch size of 8) to ensure a fair comparison. For the LLM-based evaluators, we employ open-source models with instruction templates adapted from GPTEval3D \citep{wu2024gpt4v}, as described in Appendix \ref{app:LLM-based_evaluator}. Note that we averaged the scores of multiple rendered viewpoints for these metrics. The results of our proposed T23D-CompBench are reported in Table \ref{tab:overall_performance1}, while the performance of the other three benchmarks is summarized in Table \ref{tab:overall_performance2}. The training cost and inference latency for all metrics are provided in Appendix \ref{app:training_cost_and_inference_latency}.

\begin{figure}[t]
  \centering
   \includegraphics[width=0.98\linewidth]{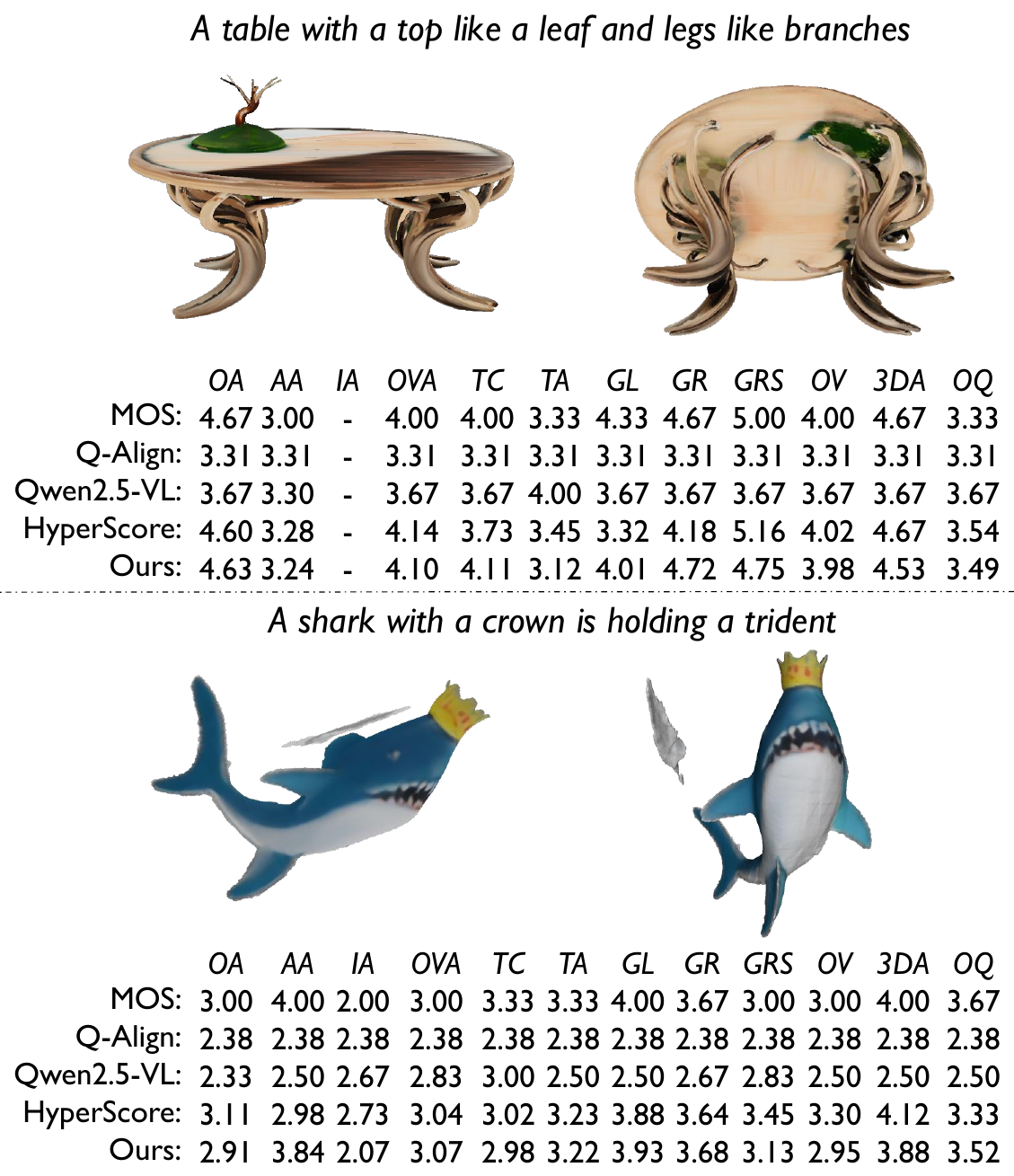}
   \setlength{\abovecaptionskip}{4pt}
   \caption{Samples with their MOSs and the predicted scores of different metrics.}
   \label{fig:gt_pre_example}
\vspace{-0.4cm}
\end{figure}

From Table \ref{tab:overall_performance1} and Table \ref{tab:overall_performance2}, we see that:
i) The proposed metric achieves consistently superior performance across all evaluation dimensions, underscoring its effectiveness for fine-grained quality assessment.
ii) For most metrics, texture quality is predicted more accurately than geometric quality. A plausible reason is that most models are trained primarily based on multi-view projections or images, which emphasize texture information while paying relatively less attention to geometric structure.
iii) LLM-based metrics demonstrate the most diverse performance range, from 0.361 (LLaVA-NeXT) to 0.575 (Qwen2.5-VL). Notably, these metrics were not explicitly fine-tuned for quality assessment tasks. The variability in their performance highlights their design focus on general-purpose visual understanding rather than specialized evaluation. Nevertheless, their strong semantic reasoning ability suggests promising potential for adaptation to quality assessment in the future.
iv) Among the fine-tuning-based metrics, ImageReward, which adopts a ranking-based pairwise learning strategy, and HyperScore, which leverages a hypernetwork to fuse different conditions, both achieve competitive performance. On T23D-CompBench, ImageReward attains an average SRCC of 0.777, while HyperScore reaches 0.809, ranking just below our proposed metric.

\begin{figure*}[t]
  \centering
   \includegraphics[width=0.88\linewidth]{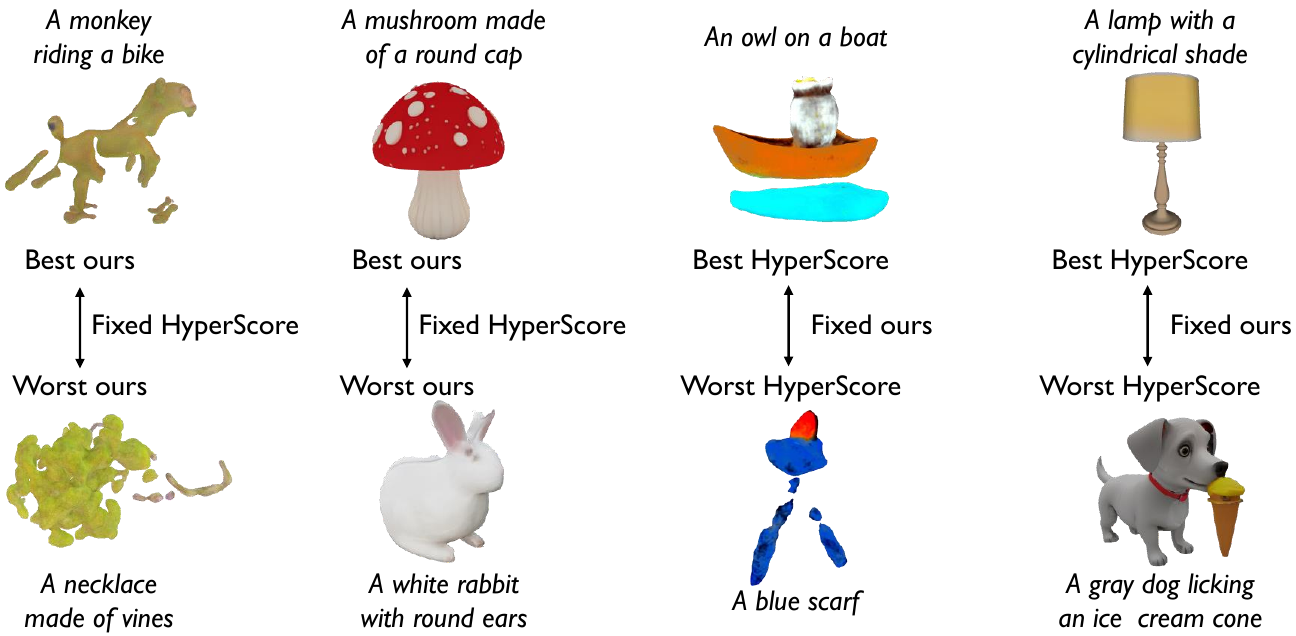}
   \setlength{\abovecaptionskip}{4pt}
   \caption{gMAD competition results between HyperScore and the proposed metric. First row: Fixed HyperScore at the low-quality level. Second row: Fixed HyperScore at the high-quality level. Third row: Fixed ours at the low-quality level. Last row: Fixed ours at the high-quality level.}
   \label{fig:gMAD_example}
\vspace{-0.2cm}
\end{figure*}

\subsubsection{Qualitative Results}

To further illustrate the advantages of Rank2Score, we conduct a qualitative analysis on representative examples in Fig. \ref{fig:gt_pre_example}. The results reveal substantial variation across different quality dimensions, underscoring the limitation of zero-shot single-score evaluators (\textit{e.g.}, Q-Align) in capturing fine-grained perceptual differences. LLM-based evaluators, such as Qwen2.5-VL, demonstrate relatively strong performance on alignment-related dimensions due to their semantic reasoning ability, but perform poorly in geometry and texture assessment, often assigning similar scores irrespective of perceptual differences. Moreover, HyperScore suffers from insufficient feature robustness, leading to inconsistent predictions across dimensions (\textit{e.g.}, for the first sample, the geometry is intact, yet the predicted score is only 3.32) and thereby reducing its reliability. In contrast, the proposed metric consistently provides stable and discriminative predictions, yielding accurate rankings across multiple dimensions. This strong alignment with human perception provides compelling evidence of its effectiveness. To provide a balanced analysis, we also present failure cases in Appendix~\ref{app:failure_case_analysis}, where Rank2Score fails to align with human judgments, offering further insight into its current limitations.

\begin{figure}[t]
    \centering
    \includegraphics[width=0.98\linewidth]{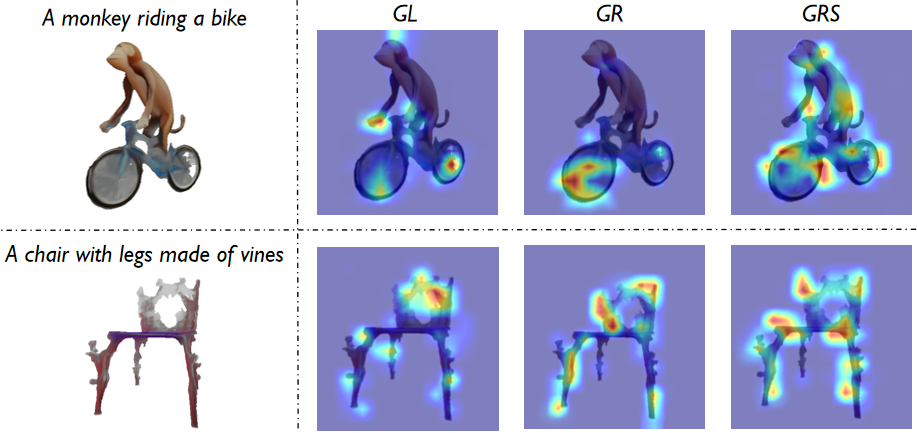}
    \caption{Visual explanations generated by XGrad-CAM for different evaluation dimensions.}
    \label{fig:xgradcam_1}
    \vspace{-0.5cm}
\end{figure}

Additionally, to further validate Rank2Score’s ability to capture geometry distortions, we conduct a qualitative analysis using XGrad-CAM \citep{fu2020axiom}. As illustrated in Fig.~\ref{fig:xgradcam_1}, Rank2Score exhibits distinct attention patterns across different geometry-related dimensions, suggesting that it can capture fine-grained geometric failures rather than relying primarily on global semantic regions. Specifically, for \textit{GL}, the model mainly attends to incomplete or missing regions, such as holes and structurally deficient areas. For \textit{GR}, the attention is concentrated on redundant structures, including floating components and irregular boundaries around the object silhouette. For \textit{GRS}, the model focuses more on surface and contour regions, which is consistent with local surface irregularities.

We also adopt the gMAD competition protocol \citep{ma2016gMAD} to perform qualitative comparisons between metrics. Using the proposed benchmark, we evaluate our metric against HyperScore, with both trained on MATE-3D benchmark. As shown in Fig. \ref{fig:gMAD_example}, HyperScore yields inconsistent predictions for high-quality samples, whereas Rank2Score more effectively captures perceptual differences across quality levels. In the third column, both samples suffer from semantic misalignment and geometric loss, yet HyperScore erroneously assigns higher quality to the top sample. In contrast, Rank2Score correctly identifies both as low quality. Similar observations in the fourth column further demonstrate the robustness of Rank2Score.

To validate the evaluation ability for ground-truth data, we download 10,000 meshes from Objaverse \citep{deitke2023objaverse} and apply the proposed Rank2Score to assess their quality. The average scores for 12 dimensions are [4.64,4.53,\\4.50,4.49,4.53,4.12,4.44,4.37,4.63,4.42, 4.56,4.25], which is generally higher than the 3D generation quality (5 denotes the highest score). Fig. \ref{fig:objarverse_example} illustrates several samples from Objaverse along with their corresponding quality scores. As the prompts do not involve attribute- and interaction-related descriptions, \textit{AA} and \textit{IA} are marked as ``$-$").

\begin{table*}[t]
\centering
\caption{Evaluation of cross-benchmark generalization. The training and test are all performed on the complete benchmark. Baseline represents the best-performing zero-shot metric on each benchmark. Results of average SRCC are reported.}
\setlength{\abovecaptionskip}{0pt}
\label{tab:cross_database_generalization}
\resizebox{\textwidth}{!}{
\begin{tabular}{c|ccc|ccc|ccc|ccc}
\toprule
Train on & \multicolumn{3}{c|}{proposed} & \multicolumn{3}{c|}{3DGCQA} & \multicolumn{3}{c|}{AIGC-T23DAQA} & \multicolumn{3}{c}{MATE-3D} \\ \midrule
Test on & 3DGCQA & \makecell{AIGC-\\T23DAQA} & MATE-3D & proposed & \makecell{AIGC-\\T23DAQA} & MATE-3D & proposed & 3DGCQA & MATE-3D & proposed & 3DGCQA & \makecell{AIGC-\\T23DAQA}
\\ \midrule
Baseline& \underline{0.489}  &0.596 & 0.619  & 0.575 & 0.596  &\textbf{0.619} &  0.575 & \textbf{0.489}& \textbf{0.619}  & 0.575&0.489   & 0.596
\\
ResNet50& 0.308  &0.368 & 0.554  & 0.065& 0.342  &0.104 &  0.331 & 0.335& 0.215  & 0.549&0.346   & 0.333
\\
ViT-B& 0.284  &0.258 &0.535   &0.201 &  0.134 & 0.086&  0.069 & 0.258& 0.027  &0.559 & 0.337  & 0.157
\\
SwinT-B& 0.318  &0.429 & 0.522  &0.218 &  0.130 & 0.121& 0.102  &0.279 &0.050   & 0.387&  0.323 & 0.045
\\
DINO v2&  0.346 &0.461 & 0.611  &0.142 & 0.168  &0.148 &0.400   &0.340 & 0.275  & 0.715&0.485   &0.356 
\\
ImageReward& 0.478  &\underline{0.643} & 0.650  & \underline{0.629} & \textbf{0.653}  & 0.575&  \textbf{0.643} & 0.438& 0.526  &0.604 & \underline{0.502}  & \textbf{0.623}
\\
HyperScore& 0.419  & 0.440 &\underline{0.661}   &0.332 & 0.268  &0.294 &  0.540 & 0.404& 0.398  & \underline{0.735}& 0.458  & 0.442
\\
\textbf{Rank2Score}&\textbf{0.530}  &\textbf{0.684} & \textbf{0.697} & \textbf{0.645}& \underline{0.621}  & \underline{0.586} & \underline{0.631} & \underline{0.449} & \underline{0.581} &\textbf{0.751} & \textbf{0.635}  & \underline{0.618}
\\ \bottomrule
\end{tabular}}
\end{table*}

\begin{figure}[t]
  \centering
   \includegraphics[width=\linewidth]{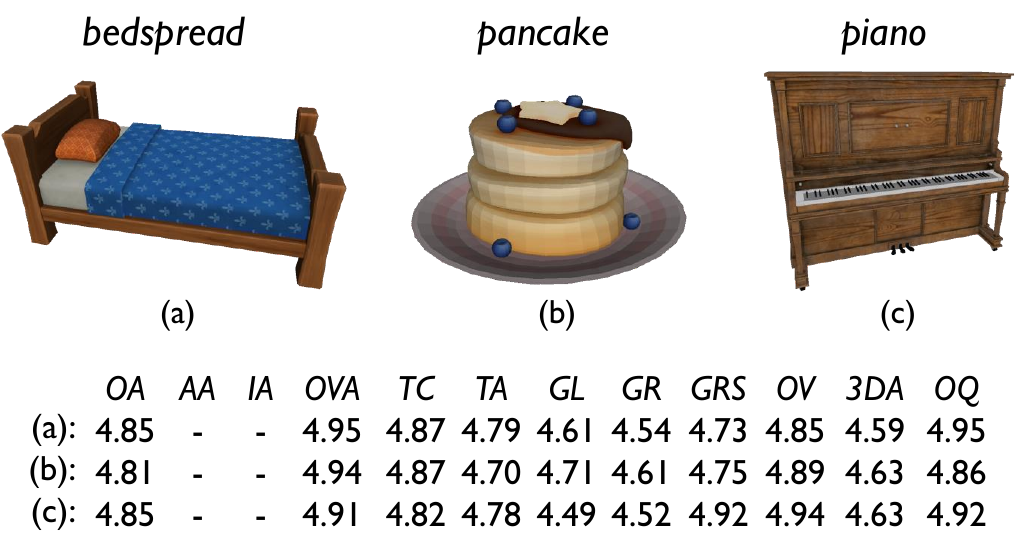}
   \setlength{\abovecaptionskip}{2pt}
   \caption{Evaluation results for examples in Objaverse.}
   \label{fig:objarverse_example}
\vspace{-0.4cm}
\end{figure}
\subsubsection{Cross-Benchmark Evaluation}
To rigorously assess the generalization capability of the proposed metric, we conduct cross-benchmark evaluations, where all metrics are trained on one benchmark and subsequently tested on the remaining benchmarks. Since the evaluation dimensions vary across benchmarks, we report the average SRCC only on the overlapping dimensions (\textit{e.g.}, when transferring from 3DGCQA to MATE-3D, only \textit{OVA} and \textit{OQ} are considered). For reference, we define the baseline as the best-performing zero-shot metric on each benchmark.

As shown in Table~\ref{tab:cross_database_generalization}, many cross-dataset SRCC values fall in the range of 0.6--0.7, highlighting the inherent difficulty of cross-dataset evaluation for T23DQA. This difficulty mainly arises from two aspects. First, existing benchmarks remain imperfect, as they are often limited in scale and lack sufficiently fine-grained evaluation dimensions. This limitation also motivates our construction of T23D-CompBench. Consequently, performance on these benchmarks may not fully reflect the intrinsic capability of an evaluation metric. Second, existing benchmarks differ substantially in prompt distribution, scoring criteria, and presentation format. For example, some benchmarks ask annotators to score rendered videos of 3D objects, while others rely on direct inspection of 3D objects, which may shift human preference labels. Despite these challenges, Table~\ref{tab:cross_database_generalization} provides empirical evidence for the cross-benchmark generalization ability of Rank2Score. When evaluated on T23D-CompBench after being trained on 3DGCQA, AIGC-T23DAQA, and MATE-3D, Rank2Score achieves SRCC values of 0.645, 0.631, and 0.751, respectively, outperforming all compared metrics in these cross-benchmark settings. These results demonstrate its relative robustness and generalization capability. Notably, ImageReward, which is also based on rank-learning, achieves competitive cross-benchmark performance.

For future work, cross-dataset generalization can be further improved in several directions. Scaling up training data with more diverse and balanced distributions can improve coverage of domain variations. Incorporating domain generalization or adaptation techniques may further enhance robustness. In addition, improving multi-modal alignment between text, 3D representations, and their 2D projections may also strengthen transferability.

\begin{table*}[t]
  \centering
  \setlength{\abovecaptionskip}{4pt}
  \caption{Ablation study of the learning strategy on T23D-CompBench. `\Checkmark' or `\XSolidBrush' means the setting is preserved or discarded. PN denotes prompt number, SD denotes score difference, and DC denotes dimension consistency. Index (1) corresponds to the baseline setting without any curriculum learning, where all training pairs are generated using the same prompt.}
  \resizebox{\linewidth}{!}{
    \begin{tabular}{c|ccc|ccccccccccccc}
    \toprule
     Index & PN & SD & DC   &AP &OA & AA & IA & OVA & TC & TA & GL & GR & GRS & OV & 3DA  & OQ\\
    \midrule
    (1) & \XSolidBrush & \XSolidBrush & \XSolidBrush &0.806 &0.805 &0.676 &0.731 &0.765 &0.880  &0.880  &0.804 	&0.789 &0.844 &0.891 & 0.757& 0.853 \\ \midrule
    (2) & \Checkmark & \XSolidBrush & \XSolidBrush &0.838 & 0.837&0.729 &0.781 &0.791 & 0.895 &0.896 	&0.821 & 0.820 	&0.851 &0.927 &0.796  &0.908\\
     (3) & \XSolidBrush & \Checkmark & \XSolidBrush  &0.826 &0.826 &0.697 &0.772 &0.783 &0.890 &0.898 &0.816  	&0.815 &0.840 &0.915 &0.773 &0.887\\
     (4) & \XSolidBrush  & \XSolidBrush & \Checkmark & 0.825 & 0.822 & 0.713 &0.774 & 0.784 &0.887  &0.891 &0.808  	&0.811 &0.832 & 0.911& 0.778 &0.884\\ \midrule
     (5) & \Checkmark & \Checkmark & \XSolidBrush  &\underline{0.849}&\underline{0.850}&\underline{0.748}&\underline{0.790}&\underline{0.813}&\underline{0.907}&\underline{0.908}&\underline{0.837}&0.826&\underline{0.852}&\underline{0.932}&\underline{0.811}&\underline{0.914}\\
    (6) & \Checkmark & \XSolidBrush & \Checkmark &0.843 &0.842&0.741&0.786&0.804&0.901&0.902&0.828&\underline{0.826}&0.847&0.927&0.800&0.911\\ 
    (7) & \XSolidBrush & \Checkmark & \Checkmark  &0.835 &0.834 &0.725&0.782&0.794&0.897&0.900&0.822&0.820&0.845&0.921&0.788&0.893\\ \midrule
    (8) & \Checkmark & \Checkmark & \Checkmark &\textbf{0.858} &\textbf{0.860} & \textbf{0.763} &\textbf{0.796} & \textbf{0.829} & \textbf{0.917} & \textbf{0.917} & \textbf{0.847} &\textbf{0.835}  	&\textbf{0.855} &\textbf{0.939} &\textbf{0.821}  &\textbf{0.920} \\
    \bottomrule
    \end{tabular}}%
  \label{tab:new_learning_strategy}%
\end{table*}%

\subsection{Ablation Study}

\subsubsection{Ablation study of the Learning Strategy}
We conduct experiments to validate the effectiveness of the three proposed learning strategies, with results reported in Table \ref{tab:new_learning_strategy}. Given that the curriculum learning strategies are specifically designed for the first stage, we evaluate the performance following the first stage training. Specifically, we evaluate both the individual and joint contributions of the strategies. Index (1) corresponds to the baseline setting without any curriculum learning, where all training pairs are constructed using the same prompt. From the table, we have the following observations: 
i) Among the three individual strategies, PN yields the most significant performance improvements across most dimensions. SD yields relatively stronger improvements on dimensions such as \textit{TC} and \textit{TA}, while DC provides more moderate but consistent improvements across most dimensions. These findings suggest that both strategies are effective. 
ii) Using two strategies generally yields better results than using any single strategy, further demonstrating that these curriculum learning strategies are complementary. Among the two-strategy settings, the combination of PN and SD achieves the best overall performance, while the other combinations also provide consistent improvements. This suggests that the three strategies contribute from different perspectives and reinforce each other when combined. 
iii) Combining all three strategies consistently achieves the best performance across all evaluation dimensions, indicating that PN, SD, and DC are complementary rather than redundant. By contributing from different perspectives, they jointly enhance the robustness and overall effectiveness of curriculum learning in Rank2Score.

\begin{table}[t]
\centering
\caption{Performance comparison between the baseline and our metric under different pair construction settings on T23D-CompBench. Average KRCC is reported.}
\label{tab:curriculum_learning_motivation}
\setlength{\abovecaptionskip}{2pt}
\begin{tabularx}{\columnwidth}{c|c|c|c|c}
\toprule
Index & Aspect & Setting & Baseline & \textbf{Ours}\\ \midrule
(1) & \multirow{2}{*}{\makecell{Prompt\\Category}}&Same prompt& 0.779& \textbf{0.790}\\
(2)& & Different prompts & 0.675 &\textbf{0.729}\\ \midrule
(3)&\multirow{4}{*}{\makecell{Score\\Difference}} & $\left |S_i-S_j  \right |  \in [0,1)$ & 0.214&\textbf{0.347} \\
(4) & 
 &$\left |S_i-S_j  \right |  \in [1,2)$ & 0.680& \textbf{0.728}\\
(5)& 
 &$\left |S_i-S_j  \right |  \in [2,3)$ & 0.861 &\textbf{0.887}\\
(6)& 
 & $\left |S_i-S_j  \right |  \in [3,4]$ & 0.895&\textbf{0.906} \\ \midrule
(7) &\multirow{2}{*}{\makecell{Dimension\\Consistency}} 
 & $c(i,j) \in [0, 0.5)$ & 0.709& \textbf{0.745}\\
(8)& 
 &$c(i,j) \in [0.5, 1.0]$ & 0.768  &\textbf{0.784}\\ 
\bottomrule
\end{tabularx}
\vspace{-0.3cm}
\end{table}
To better illustrate the limitations of the baseline, we present a comparison of performance between the baseline and our metric under various pair construction settings in Table \ref{tab:curriculum_learning_motivation}. Observing (1)-(8), performance declines on more challenging pairs, including different-prompt pairs (\textit{e.g.}, KRCC decreasing from 0.779 to 0.675), pairs with small MOS differences (\textit{e.g.}, 0.895 for [3,4] \textit{vs.} 0.214 for [0,1)), and pairs exhibiting high variance across quality dimensions (\textit{e.g.}, KRCC decreasing from 0.768 to 0.709). These results further confirm that such pairs are more difficult to rank and can be regarded as hard samples. Comparing (7)-(8) with (1)-(2) and (3)-(6), dimension consistency exerts a relatively smaller influence on ranking accuracy compared to prompt category and score difference. Moreover, results for the baseline and our metric further demonstrate that the proposed curriculum learning strategies are effective in handling hard samples.

\subsubsection{Ablation study of the Loss Function} 
The proposed network is trained using the rank loss $L_{rank}$, the supervised constructive loss $L_{cons}$, and the MSE loss $L_{mse}$. We explore how the three loss functions contribute to our metric. From Table \ref{tab:ablation_loss_function}, it can be observed that using only $L_{rank}$ already yields strong performance, while incorporating $L_{cons}$ further improves evaluation across all dimensions. Moreover, in the second stage, applying $L_{mse}$ to align the predicted scores with the MOS further enhances performance.

\begin{table*}[t]
\centering
\caption{Ablation study of the loss function on T23D-CompBench. Results of SRCC are reported.}
\label{tab:ablation_loss_function}
\resizebox{\textwidth}{!}{
\begin{tabular}{ccc|ccccccccccccc}
\toprule
$L_{rank}$ & $L_{cons}$ & $L_{mse}$ & AP & OA & AA & IA & OVA & TC & TA & GL & GR & GRS & OV & 3DA  & OQ  \\ \midrule

\Checkmark & \XSolidBrush & \XSolidBrush & 0.833 & 0.837	&0.678 	&0.773 	&0.799 	&0.888 	&0.892 	& 0.850	&0.817 	& 0.868	&0.910 	&0.811 	&0.878 
\\
\Checkmark & \Checkmark & \XSolidBrush &0.858 &0.860 & 0.763 	& 0.796	& 0.829 	& \underline{0.917} & \underline{0.917}	& 0.847 	& 0.835 	&0.855 & \underline{0.939}& 0.821 &\underline{0.920} 
\\
\Checkmark & \XSolidBrush & \Checkmark &\underline{0.890}  &\underline{0.897} 	&\underline{0.810} 	&\underline{0.892} 	&\underline{0.911} 	& 0.903	& 0.925	& \textbf{0.902}	& \underline{0.870}	& \textbf{0.902}	& 0.920	& \underline{0.849}	&0.895
\\
\Checkmark & \Checkmark & \Checkmark &\textbf{0.908 }&\textbf{0.912} & \textbf{0.880} 	&\textbf{0.907} 	&\textbf{0.928} 	& \textbf{0.920} &\textbf{0.944} 	&\underline{0.898} &\textbf{0.878}  	&\underline{0.887} &\textbf{0.948} &\textbf{0.862}  &\textbf{0.934}
\\ \bottomrule
\end{tabular}}
\vspace{-0.2cm}
\end{table*}

\begin{table*}[t]
\centering
\caption{Ablation study of the level prompt design on T23D-CompBench. Results of SRCC are reported.}
\label{tab:ablation_prompt_design}
\resizebox{\textwidth}{!}{
\begin{tabular}{cc|ccccccccccccc}
\toprule
Type & Position & AP & OA & AA & IA & OVA & TC & TA & GL & GR & GRS & OV & 3DA  & OQ  \\ \midrule
Regression & - &0.875  &0.872 	&0.851 	&0.892 	& 0.892	& 0.891	& 0.911	& 0.869	&0.842 	& 0.864	& 0.917	&0.815 	&0.889 
\\ \midrule
Fixed & - & 0.899 & 0.900	& 0.810	& 0.904	& 0.923	& 0.916	& \underline{0.945}	&\underline{0.901} 	&0.874 	&\underline{0.894} 	&0.949 	&0.846 	&0.920 
\\ \midrule
& begin &0.904 	&0.904 	& 0.857	& \underline{0.918}	& \underline{0.924}	&0.916 	&0.943 	&\textbf{0.901} 	&0.873 	&\textbf{0.896} 	& \underline{0.949}	& 0.845 & 0.919
\\
Learnable & middle &\textbf{0.908} &\textbf{0.912} & \textbf{0.880} 	&0.907	&\textbf{0.928} 	& \textbf{0.920} &0.944	&0.898 &\underline{0.878}  	&0.887 &0.948 &\textbf{0.862}  &\textbf{0.934}
\\
& end & \underline{0.906}	&\underline{0.908}	&\underline{0.861} 	&\textbf{0.921} 	&0.924 	& \underline{0.919}	& \textbf{0.945}	& 0.896	& \textbf{0.879}	& 0.893	& \textbf{0.949}	& \underline{0.850} & \underline{0.927}
\\ \bottomrule
\end{tabular}}
\vspace{-0.2cm}
\end{table*}

\begin{table*}[t]
\centering
\caption{Ablation study of the aggregation strategy between weighted visual and textual features on T23D-CompBench. Results of SRCC are reported.}
\label{tab:ablation_aggregation_strategy}
\resizebox{\textwidth}{!}{
\begin{tabular}{c|ccccccccccccc}
\toprule
Aggregation & AP & OA & AA & IA & OVA & TC & TA & GL & GR & GRS & OV & 3DA  & OQ  \\ \midrule

$\widetilde{F_I^d} + \widetilde{F_P^d}$ & 0.886 &0.879 	&\textbf{0.890} &\underline{0.889}	&0.898 	& 0.902	& 0.921	& 0.873	& 0.843	&\underline{0.873} 	&0.921 	&0.828 	&0.917 
\\
$\widetilde{F_I^d} \odot \widetilde{F_P^d}$ & \underline{0.893} &\underline{0.890} 	&\underline{0.885} 	&0.867 	&\underline{0.909} 	&\underline{0.910} 	& \underline{0.930}	& \underline{0.886}	& \underline{0.857}	&0.870 	&\underline{0.934} 	&\underline{0.847} 	& \underline{0.927}
\\
$\widetilde{F_I^d} \oplus \widetilde{F_P^d} $&\textbf{0.908 }&\textbf{0.912} & 0.880 	&\textbf{0.907} 	&\textbf{0.928} 	& \textbf{0.920} &\textbf{0.944} 	&\textbf{0.898} &\textbf{0.878}  	&\textbf{0.887} &\textbf{0.948} &\textbf{0.862}  &\textbf{0.934}
\\ \bottomrule
\end{tabular}}
\vspace{-0.3cm}
\end{table*}

\subsubsection{Ablation study of the Level Prompt Design} To assess the contribution of level features, we first compare our metric with a baseline that directly applies MLP regression to the fused features. As shown in Table \ref{tab:ablation_prompt_design}, directly applying MLP regression performs the worst, indicating that quality assessment is more effectively modeled with discrete levels, consistent with human scoring behavior.

To further assess the effectiveness of learnable level prompts, we conduct experiments with fixed prompts of the form ``The quality of this image is \textit{[quality level]} \textit{[dimension name]}". We also investigate different positions of ``\textit{[quality level]} \textit{[dimension name]}". The results indicate that appropriately designed learnable prompts consistently outperform fixed prompts, thereby reducing the dependence on handcrafted prompt design. Notably, placing the token in the middle position yields the best performance, as bidirectional contextual information enables a more precise semantic interpretation of the class token.

\begin{figure}[t]
  \centering
   \includegraphics[width=0.94\linewidth]{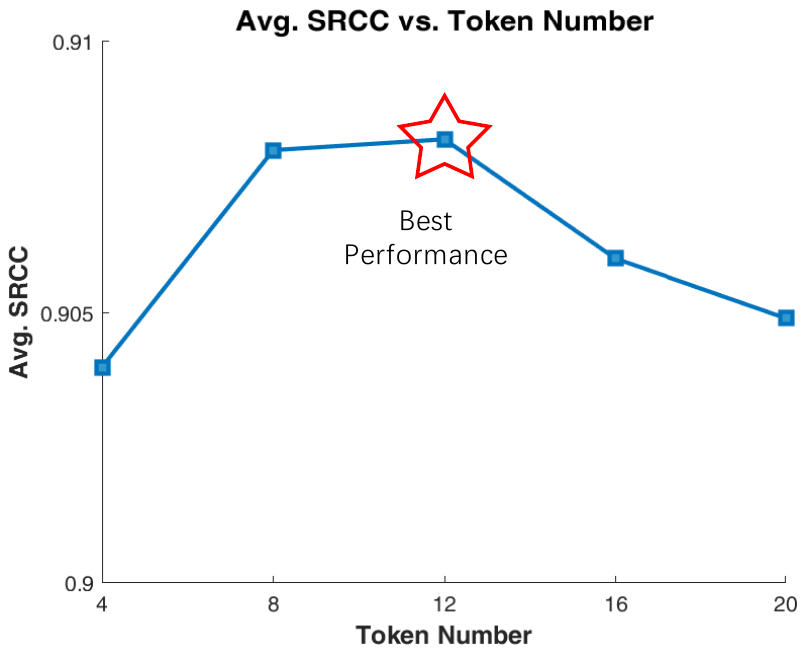}
   \setlength{\abovecaptionskip}{0pt}
   \caption{Performance comparison of the learnable token
length on T23D-CompBench.}
   \label{fig:token_number}
\vspace{-0.4cm}
\end{figure}
Building on the observation that learnable prompts yield superior performance, we further investigate the effect of learnable token length $K$ on evaluation performance. As shown in Fig. \ref{fig:token_number}, the optimal performance is achieved when $K = 12$. Short prompts risk omitting essential information due to limited expressiveness, while overly long prompts may introduce irrelevant content that obscures salient cues. These results highlight the importance of balancing learnable token length to preserve informativeness while minimizing noise.

\begin{table*}[t]
\centering
\setlength{\abovecaptionskip}{2pt}
\caption{Ablation study of the number of rendered viewpoints on T23D-CompBench. Results of SRCC are reported.}
\label{tab:ablation_viewpoint_count}
\resizebox{\textwidth}{!}{
\begin{tabular}{c|ccccccccccccc}
\toprule
$N_v$ & AP & OA & AA & IA & OVA & TC & TA & GL & GR & GRS & OV & 3DA  & OQ  \\ \midrule

4&0.895  &0.893 	& 0.872	&0.875 	&0.911 	&0.910 	&0.934 	&0.895 	&0.873 	&0.876 	&0.935 	&0.848 	&0.917 
\\
6 &\textbf{0.908 }&\textbf{0.912} & \textbf{0.880} 	&\textbf{0.907} 	&\textbf{0.928} 	& \textbf{0.920} &\underline{0.944} 	&0.898 &\underline{0.878}  	&\underline{0.887} &\textbf{0.948} &\underline{0.862}  &\textbf{0.934}
\\
9 &\underline{0.906}  &\underline{0.905} 	&\underline{0.878} 	&\underline{0.900} 	&\underline{0.925} 	&\underline{0.918} 	& \textbf{0.944}	&0.898 	&\textbf{0.880} 	&\textbf{0.890} 	&\underline{0.945} 	& 0.856	&\underline{0.933} 
\\
12 & 0.904 & 0.905	& 0.868	&0.893 	&0.918 	&0.917 	&0.941 	&\underline{0.907} 	& 0.874	& 0.883	& 0.945	& \textbf{0.864}	&0.932 
\\
16 & 0.903 & 0.905	&0.876 	&0.892 	&0.921 	&0.910 	&0.936 	&\textbf{0.930} 	& 0.874	&0.884 	& 0.940	&0.858 	&0.932 
\\ \bottomrule
\end{tabular}}
\vspace{-0.2cm}
\end{table*}

\begin{table*}[h]
\centering
\setlength{\abovecaptionskip}{2pt}
\caption{Ablation study of input modalities and fusion strategies on T23D-CompBench. Results of SRCC are reported.}
\label{tab:normal_map}
\resizebox{\textwidth}{!}{
\begin{tabular}{c|c|ccccccccccccc}
\toprule
Modality & Fusion Strategy & AP & OA & AA & IA & OVA & TC & TA & GL & GR & GRS & OV & 3DA  & OQ  \\ \midrule

Texture only&-&\underline{0.908 }&\underline{0.912} & \underline{0.880} 	&\textbf{0.907} 	& 0.928 & \underline{0.920} &\textbf{0.944} 	&0.898 &0.878 	&0.887 &\underline{0.948} &\underline{0.862}  &\underline{0.934}
\\
Normal only&-&0.887  &0.889 	&0.826 	& 0.868	& 0.915	& 0.901	& 0.918	& \underline{0.919}	& 0.877	& 0.880	& 0.917	&0.855 	&0.882 
\\ \midrule
&Average & 0.896 & 0.900	&0.850 	&0.873 	&\underline{0.929} 	&0.914	&0.931	&0.919	&0.874 	&0.888 	&0.925 	&0.857 	&0.887
\\ 
Texture + Normal& CA& 0.904& 0.898 &0.875	&0.885 	&0.919  &\textbf{0.933}  &0.938  &0.911  & \underline{0.882} &\underline{0.893}  &0.931  & 0.859  &0.926
\\
&AdaLN&\textbf{0.916}  &\textbf{0.913}	&\textbf{0.887} & \underline{0.901} &\textbf{0.933}  &0.916  & \underline{0.942} & \textbf{0.921} & \textbf{0.899} &\textbf{0.905}  &\textbf{0.952}   &\textbf{0.878}  &\textbf{0.939}  
\\
\bottomrule
\end{tabular}}
\vspace{-0.3cm}
\end{table*}

\subsubsection{Ablation Study of the Aggregation Strategy} 
To explore the impact of different aggregation strategies for combining weighted visual and textual features, we compare three strategies: concatenation ($\oplus$, used in our metric), addition (+), and element-wise multiplication ($\odot$). As shown in Table \ref{tab:ablation_aggregation_strategy}, concatenation consistently yields the highest performance across all evaluation dimensions. In contrast, addition and multiplication yield lower performance, confirming concatenation as the most effective aggregation strategy.

\subsubsection{Ablation study of the Viewpoint Count}
We render each textured mesh into $N_v = 6$ images from six orthogonal viewpoints (\textit{i.e.}, along the positive and negative directions of the x, y, and z axes) for evaluation. To investigate the influence of viewpoint count, we further evaluate the proposed metric under varying $N_v$ values, using camera settings similar to those in \citep{zhang2025hyperscore}.
As shown in Table \ref{tab:ablation_viewpoint_count}, performance initially improves with more viewpoints, but declines when $N_v$ becomes too large. This suggests that while increasing $N_v$ provides additional visual information beneficial for prediction, excessive viewpoints may introduce redundancy and noise, negatively affecting performance. Furthermore, a higher $N_v$ also leads to increased computational cost. To balance accuracy and efficiency, we adopt $N_v = 6$ as the default setting.

\subsubsection{Ablation study of the Input modality}

To evaluate the impact of geometric information, we incorporate normal maps as an additional input to the evaluator. Specifically, we investigate different input modalities and fusion strategies, including averaging, cross-attention (CA) \citep{vaswani2017attention}, and adaptive layer normalization (AdaLN) \citep{guo2022adaln}. For CA, texture features are used as queries and normal features serve as keys and values; For AdaLN, normal features are used to generate weights an biases for texture features. The experimental results are reported in Table \ref{tab:normal_map}.

\begin{figure}
    \centering
    \includegraphics[width=0.88\linewidth]{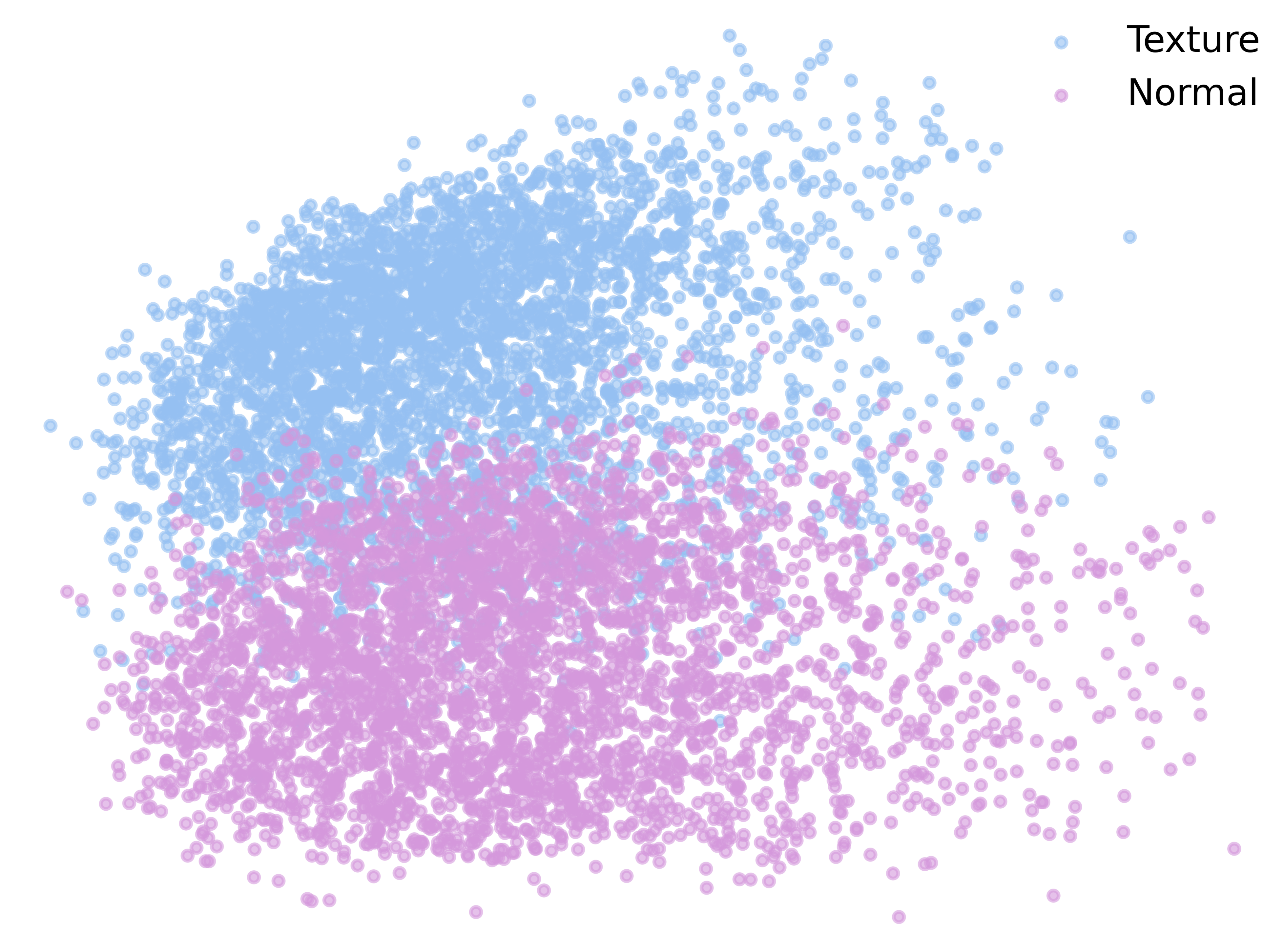}
    \caption{PCA of the texture and normal features on complete T23D-CompBench.}
    \label{fig:pca_visualization}
\end{figure}

From the table, we have the following observations: 
i) Directly averaging texture and normal features does not lead to clear overall quality improvements. 
To better understand this phenomenon, we perform Principal Component Analysis (PCA) on the texture and normal features extracted from T23D-CompBench and project them into a 2D embedding space, as shown in Fig.~\ref{fig:pca_visualization}. 
The visualization reveals a noticeable discrepancy between the two modalities in the embedding space, suggesting that simple averaging is insufficient to effectively fuse texture and normal features.
ii) Although naive averaging brings limited gains, normal features are still beneficial for geometry-related evaluation, as reflected by the improved performance on geometry-sensitive dimensions such as \textit{GL}.
iii) Adopting more effective feature fusion strategies such as CA and AdaLN can achieve better performance than direct averaging. 
In particular, AdaLN not only improves geometry-related evaluation performance but also surpasses the original texture-only setting in most metrics, indicating that normal features can provide complementary geometric cues when properly integrated. 
Therefore, investigate more effective fusion strategies for different modalities remains an interesting and important avenue for future work.

\begin{table*}[h]
\centering
\setlength{\abovecaptionskip}{2pt}
\caption{Ablation study of the rotation angles on T23D-CompBench. Results of SRCC are reported.}
\label{tab:ablation_rotation_angle}
\resizebox{\textwidth}{!}{
\begin{tabular}{c|ccccccccccccc}
\toprule
Angle & AP & OA & AA & IA & OVA & TC & TA & GL & GR & GRS & OV & 3DA  & OQ  \\ \midrule
- &\textbf{0.908 }&\textbf{0.912} & \textbf{0.880} 	&\textbf{0.907} 	&\textbf{0.928} 	& 0.920&\underline{0.944}	&0.898 &\underline{0.878} 	&0.887 &0.948 &\textbf{0.862}  &\textbf{0.934}
\\
$30^{\circ} $ & 0.903 &0.902 	&0.877 	&0.896 	&0.915 	&0.896 	& \textbf{0.945}	& \textbf{0.900}	&\textbf{0.879} 	&\textbf{0.889} 	&\underline{0.950} 	&0.855 	&\underline{0.929} 
\\
$45^{\circ} $ & \underline{0.906} &\underline{0.906} 	& \underline{0.879}	& \underline{0.904}	& \underline{0.918}	& \textbf{0.924}	& 0.944	&\underline{0.899} 	&0.876 	&\underline{0.889} 	&\textbf{0.951} 	&0.856 	&0.929 
\\
$60^{\circ} $ & 0.905 & 0.903	& 0.879	& 0.896	& 0.916	& \underline{0.924}	& 0.944	& 0.899	& 0.876	& 0.888	& 0.948	& \underline{0.857}	& 0.928
\\ \bottomrule
\end{tabular}}
\vspace{-0.3cm}
\end{table*}

\subsubsection{Ablation study of the Rotation Angle} 
An effective 3D evaluation metric should exhibit robustness to variations in scale and viewpoint. The proposed metric ensures scale invariance by normalizing all meshes into a unit sphere before rendering. In addition, it demonstrates rotation robustness due to the diverse orientations of training samples. To further verify this property, we evaluate the metric under different azimuth angles and report the results in Table \ref{tab:ablation_rotation_angle}. The performance remains relatively stable across angles, confirming the metric’s resilience to rotational changes.

\subsection{Application}

Objective evaluators are not only valuable for measuring the quality of generative models, but can also be leveraged as rewards to guide the generation process. Building upon MVC-ZigAL \citep{zhang2025refining}, we further investigate the applicability of the proposed evaluator in this setting. Specifically, we re-train MVC-ZigAL using the official open-source implementation with default weights, and incorporate three quality evaluators (\textit{i.e.}, ImageReward, HyperScore, and our proposed Rank2Score) as reward signals. To ensure a fair comparison among evaluators, we exclusively use the predicted overall quality score.

For quantitative analysis, we compare MVC-ZigAL trained with three different evaluators on the MATE-3D benchmark. For each prompt, the three models generate the corresponding 3D samples, which are presented together to human participants for direct comparison. Following ITU-R BT.500 \citep{bt500}, we conduct a user study with 16 participants, who are asked to choose the sample with the best overall quality. The win rate is then computed based on human preferences. As shown in Table \ref{tab:reward_experiment}, MVC-ZigAL optimized with our evaluator achieves the highest win rate, demonstrating the effectiveness of Rank2Score in guiding T23D generation and its potential as a reliable metric for downstream applications.

\begin{table}[t]
\centering
\caption{Generation quality comparison of MVC-ZigAL under three different reward metrics.}
\label{tab:reward_experiment}
\begin{tabularx}{\columnwidth}{c|c|c|c}
\toprule
Metric & \makecell[c]{MVC-ZigAL \\ (ImageReward)} & \makecell[c]{MVC-ZigAL \\ (HyperScore)} & \makecell[c]{MVC-ZigAL \\ (Ours)} \\ \midrule
Win Rate (\%)  & 7.23 & \underline{28.75} & \textbf{64.02} \\ \bottomrule
\end{tabularx}
\vspace{-0.2cm}
\end{table}

\begin{figure*}[t]
  \centering
   \includegraphics[width=0.94\linewidth]{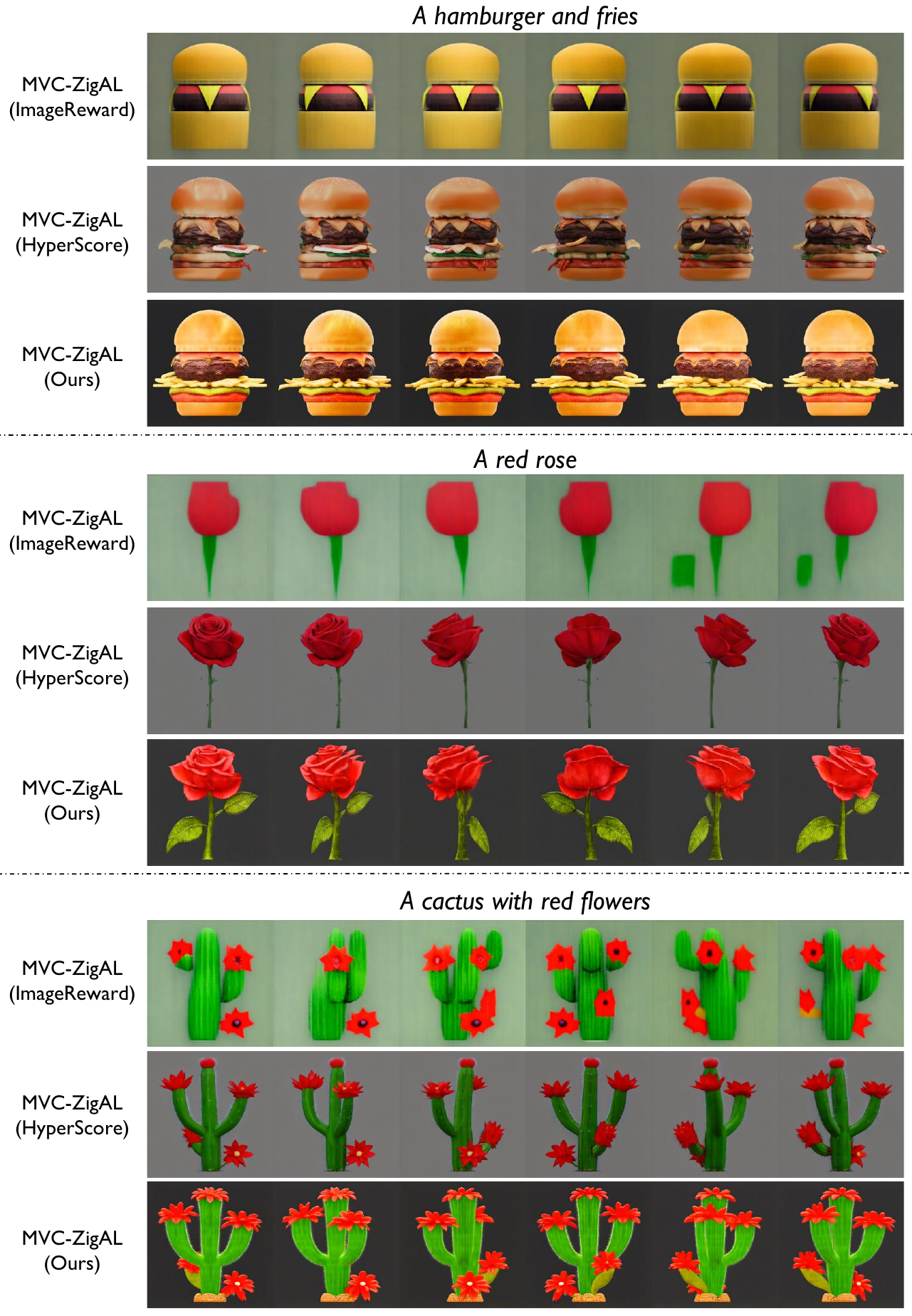}
   \caption{Text-to-multiview generation results of MVC-ZigAL trained with different reward metrics (ImageReward, HyperScore, and Ours).}
   \label{fig:application}
\vspace{-0.3cm}
\end{figure*}

In addition, we further select several representative samples for qualitative visualization. From the results in Fig. \ref{fig:application}, prompts involving multiple objects, such as \textit{``A hamburger and fries"}, remain challenging. Specifically, generations guided by ImageReward and HyperScore fail to include the \textit{``fries"} component, resulting in semantic misalignment. Moreover, subsequent examples indicate a lack of fine-grained details, for instance, roses being generated without leaves. These results suggest that although ImageReward and HyperScore yield improvements over the baseline metric, their performance is unstable across different cases. In contrast, the proposed evaluator better preserves semantic alignment and fine-grained structures, consistently producing more robust and higher-quality multiview generations, thereby demonstrating its effectiveness as a reliable reward signal for guiding the generation process.

\section{Conclusion}

In this paper, we addressed the challenge of fine-grained quality evaluation for T23D generation. To this end, we first constructed a comprehensive benchmark, T23D-CompBench, comprising 3,600 textured mesh samples annotated across twelve carefully designed evaluation dimensions. Building on this benchmark, we proposed a two-stage, ranking-based fine-grained evaluator, named Rank2Score, which leverages curriculum learning strategies to adaptively predict quality across diverse dimensions. Extensive experiments demonstrate that our metric provides a reliable and fine-grained assessment of T23D generation, offering a valuable foundation for future research and practical applications in this emerging field.

\begin{appendices}



\section{More Details on Benchmark Construction}\label{app:more_details_on_benchmark_construction}
\subsection{Instruction of Prompt Pipeline}\label{app:instruction_of_prompt_pipeline}
\begin{figure*}[t]
  \centering
   \includegraphics[width=0.98\linewidth]{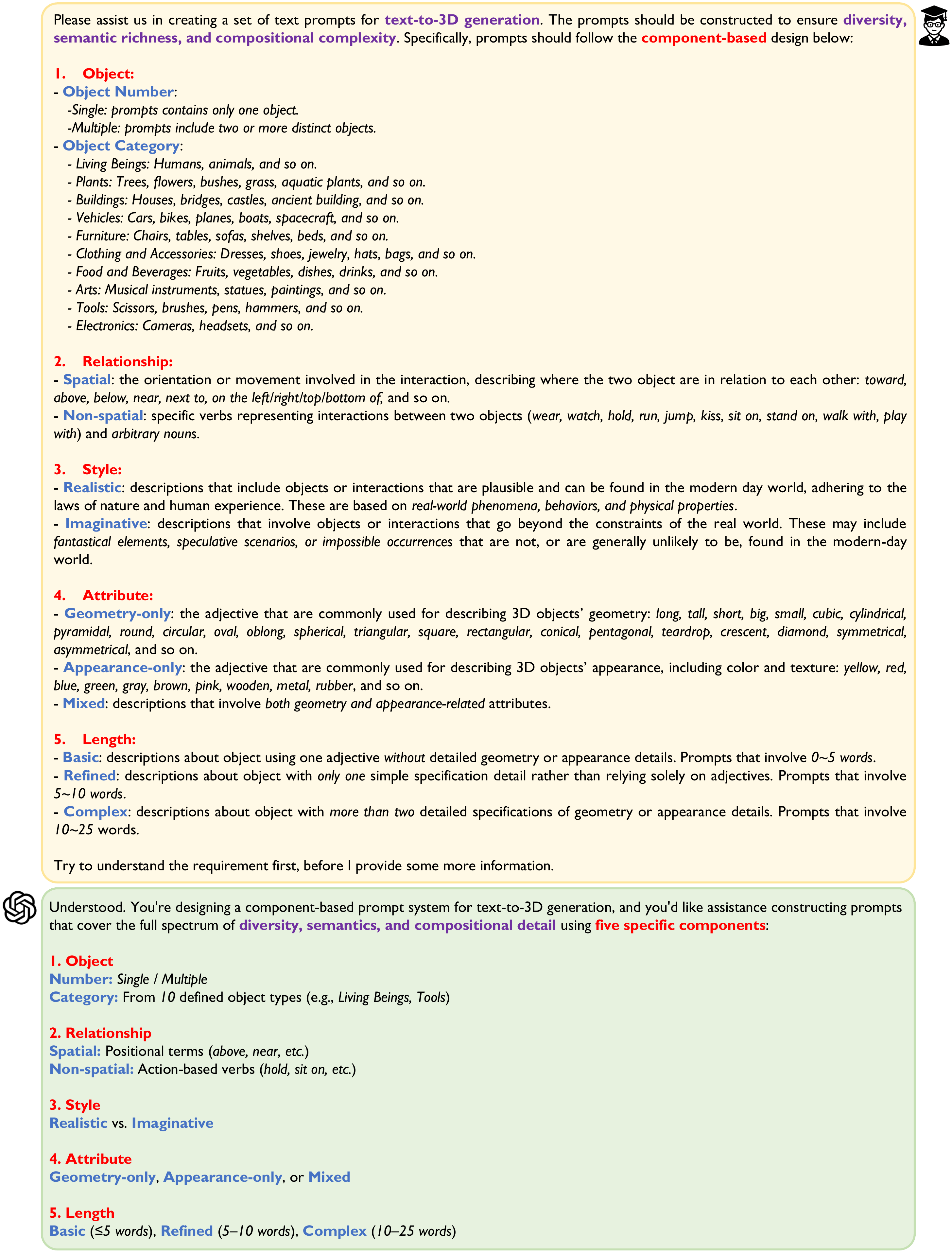}
   \caption{Examples of GPT-4o instruction templates for prompt generation. (Part 1/2).}
   \label{fig:intruction_template1}
\vspace{-0.3cm}
\end{figure*}

\begin{figure*}[t]
  \centering
   \includegraphics[width=0.98\linewidth]{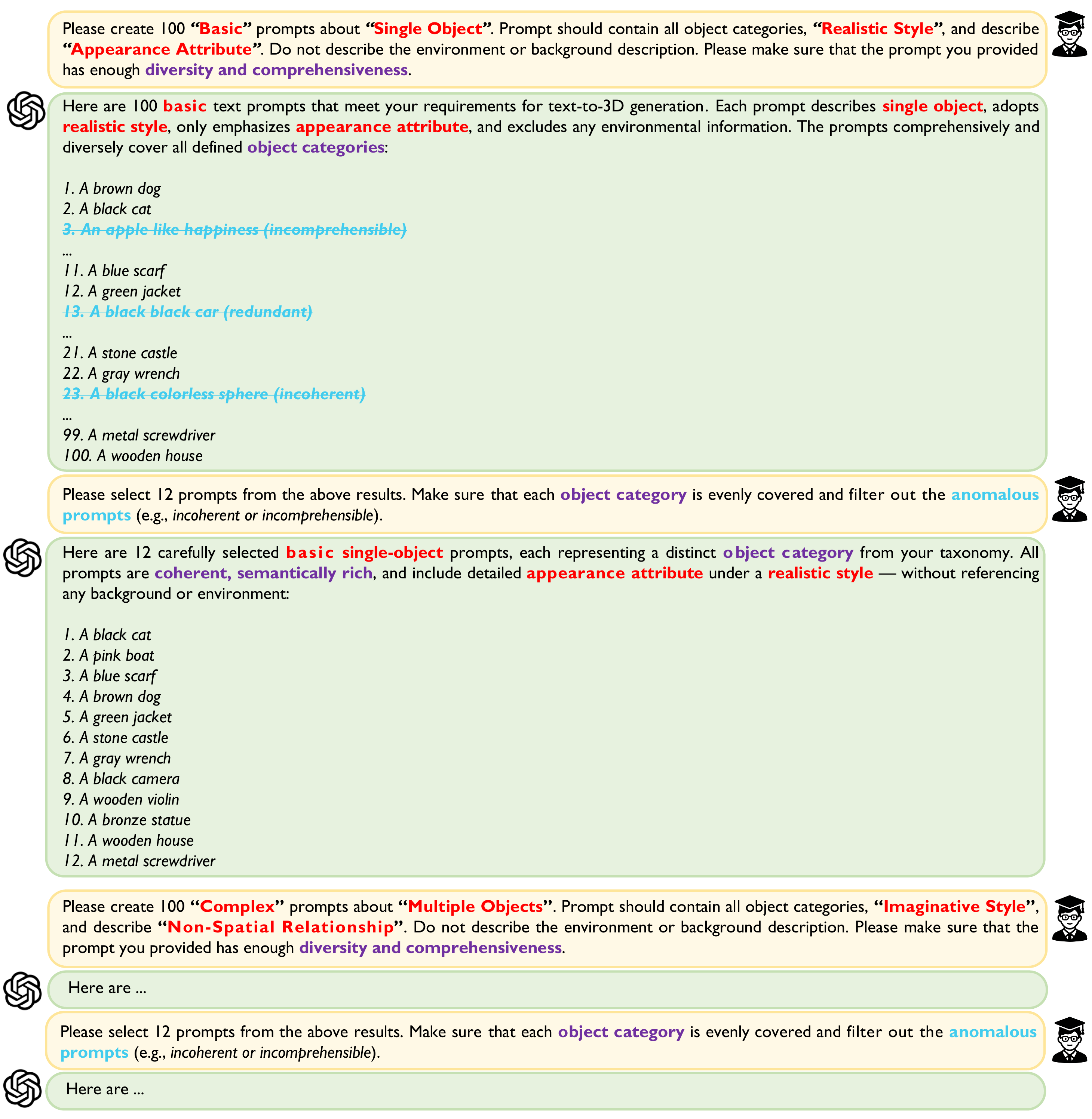}
   \caption{Examples of GPT-4o instruction templates for prompt generation. (Part 2/2).}
   \label{fig:intruction_template2}
\vspace{-0.3cm}
\end{figure*}

After defining five prompt components and twelve sub-components, we design a unified instruction template to guide GPT-4o in automatically generating text prompts. For each component combination, we first generate 100 candidate prompts and subsequently filter out anomalous ones. Since GPT-4o sometimes produces prompts containing incomprehensible concepts (\textit{e.g.}, \textit{``A sphere of unknown identity"}, \textit{``A asdf qwer lkjh zxcv"}), incoherent descriptions (\textit{e.g.}, \textit{``A metallic wooden stone"}), or redundant prompts (\textit{e.g.}, \textit{``An apple with texture texture on its surface"}), generating candidates followed by manual filtering of incoherent or incomprehensible prompts is necessary. 
\clearpage
Ultimately, we select twelve valid prompts from the 100 candidates, ensuring an even distribution across object categories. The detailed template structure is illustrated in Fig. \ref{fig:intruction_template1} and Fig. \ref{fig:intruction_template2}. This template provides GPT-4o with a clear understanding of the generation task, enabling it to produce structured and semantically coherent prompts based on the specified inputs. By modifying different component configurations, GPT-4o can flexibly generate diverse prompts for a total of 30 combinations. This structured template ensures controllability and reproducibility across different tasks.

\subsection{Definition of Different Quality Dimensions}\label{app:definition_of_different_quality_dimensions}
The quality of generated textured meshes is influenced by multiple perceptual factors. To enable fine-grained and interpretable evaluation, we define four evaluation dimensions comprising twelve sub-dimensions, each capturing a distinct aspect of human-perceived quality. The dimensions are described as follows:

\begin{itemize}
    \item \textbf{Textual-3D Alignment.} This dimension assesses the semantic consistency between the generated 3D content and the input prompt. It consists of four sub-dimensions. i)  \textbf{Object Alignment.} Evaluates whether the category, number, and identity of generated objects match the prompt.
    ii) \textbf{Attribute Alignment.} Measures the degree to which geometric attributes (\textit{e.g.}, shape, size) and appearance attributes (\textit{e.g.}, color, material) align with the prompt. iii) \textbf{Interaction Alignment.} Assesses whether the actions, spatial relations, and interactions among objects described in the prompt are accurately represented in the generated 3D textured meshes. iv) \textbf{Overall Alignment.} Provides a holistic measure of semantic alignment, integrating the above factors.
    \item \textbf{3D Visual Quality.} Evaluates the visual and structural fidelity of the 3D mesh, independent of the input prompt. It consists of six sub-dimensions. 
    i) \textbf{Texture Clarity.} Assesses the sharpness and resolution of the texture, ranging from blurry to crisp. ii) \textbf{Texture Aesthetics.} Evaluates the visual appeal of the texture, considering factors such as spotting, color smoothness, and visual clutter on the surface. 
    iii) \textbf{Geometry Loss.} Measures the extent of missing or incomplete geometry, such as holes, flattened regions, or missing components (\textit{e.g.}, a table missing a leg). iv) \textbf{Geometry Redundancy.} Identifies redundant or noisy geometry, including floating fragments or unnatural white boundary artifacts. v) \textbf{Geometry Roughness.} Assesses the surface smoothness of the 3D mesh, particularly along edges and contours. vi) \textbf{Overall Visual Quality.} Provides a comprehensive assessment of geometry and appearance quality.
    \item \textbf{3D Authentic.} Evaluates the realism and plausibility of the generated 3D object, penalizing anatomically or physically implausible structures caused by over-symmetry or generative artifacts (\textit{e.g.}, a human with three legs or a rabbit with two tails).
    \item \textbf{Overall Quality.} Provides a global judgment that integrates the dimensions of alignment, visual quality, and authenticity, reflecting the overall perceptual quality of the generated 3D asset relative to the input prompt.
\end{itemize}

\subsection{Subjective Experiment Procedure}\label{app:subjective_experiment_procedure}

\textbf{Training Session.} 
To ensure the reliability of subjective ratings, we incorporate a training session before the formal evaluation. Specifically, we select 10 representative samples whose prompts are excluded from the proposed benchmark to avoid bias or memorization effects. These samples are carefully curated to cover a broad range of quality levels across multiple dimensions, thereby helping participants develop an accurate understanding of the evaluation criteria. Expert annotators first assign reference scores to the training samples. During the training, participants are required to view all samples twice. In the first pass, they become familiar with the 3D content; in the second pass, they are asked to assign scores across all sub-dimensions. We then compute the correlation between their scores and the reference scores. Only participants whose annotations demonstrate a strong agreement with the reference scores proceed to the reliable evaluation. Conversely, if viewers assign biased scores from the references, we repeat the training procedure until they provide reasonable results. This procedure ensures that all annotators are well-calibrated and capable of providing reliable and interpretable scores throughout the subjective study.

\begin{figure}[t]
  \centering
   \includegraphics[width=\linewidth]{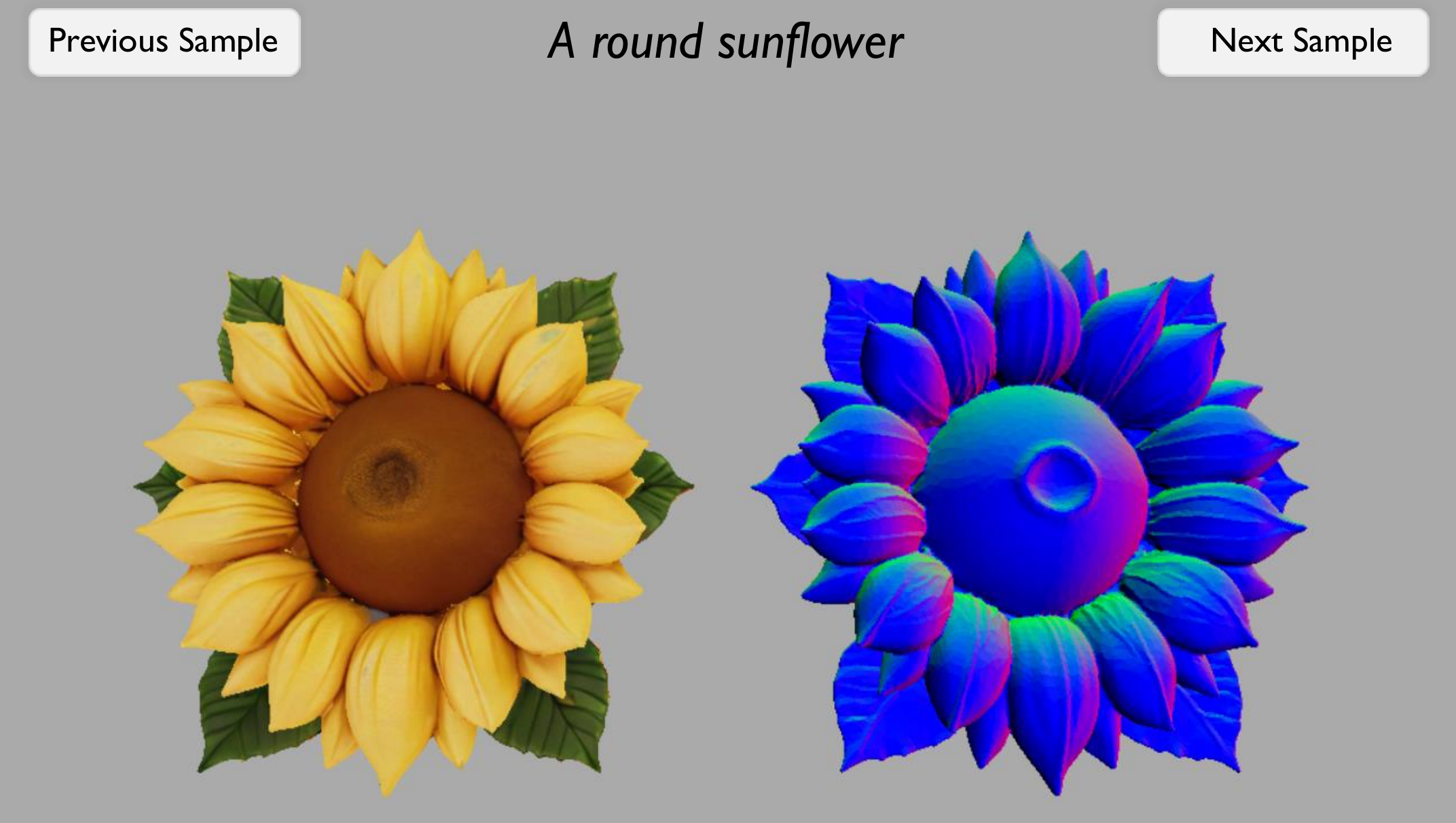}
   \setlength{\abovecaptionskip}{-2pt}
   \caption{Illustration of the environment platform.}
   \label{fig:environment_platform}
\vspace{-0.5cm}
\end{figure}

\begin{table*}[h]
\centering
\setlength{\abovecaptionskip}{2pt}
\caption{Comparison of training cost and inference latency on T23D-CompBench.}
\label{tab:cost_and_latency_reviewer1}
\resizebox{\textwidth}{!}{
\begin{tabular}{c|c|ccc|ccc}
\toprule
\multirow{3}{*}{Type} & \multirow{3}{*}{Metric} 
& \multicolumn{3}{c|}{Single-view} 
& \multicolumn{3}{c}{Six-view} \\
\cmidrule{3-5} \cmidrule{6-8}

\multicolumn{1}{c|}{} & \multicolumn{1}{c|}{} & \makecell[c]{FLOPs \\ (G)} & \makecell[c]{Training Cost \\ (ms/step)}& \makecell[c]{Inference Latency\\ (ms/sample)}
& \makecell[c]{FLOPs \\ (G)} & \makecell[c]{Training Cost \\ (ms/step)}& \makecell[c]{Inference Latency\\ (ms/sample)} \\
\midrule

& CLIPScore \citep{hessel2021clipscore} & 56.26 & - & 20.69  & 337.53 & - & 34.21
\\
&BLIPScore \citep{li2022blip} &62.66 & - &22.66  & 375.97 & - & 61.54
\\
&Aesthetic Score \citep{schuhmann2022laion} & 51.90  & - & 16.65  & 311.38 & - & 38.34 
\\
&ImageReward \citep{xu2024imagereward}  & 66.88 & - & 28.80 & 401.26 & - & 107.73
\\
&DreamReward \citep{ye2024dreamreward}  & 114.20 & - & 43.53 & 685.19 & - & 193.16 
\\
Zero-shot &HPS v2 \citep{wu2023human}  & 123.49  & - & 49.39  & 740.93 &- & 97.18
\\
&Q-Align \citep{wang2023exploring}  & 936.35 & - &  255.52 & 5618.09 &- & 1505.36
\\
& LLaVA-NeXT (8B) \citep{li2024llava}  & 32356.74 & - & 25975.97  & 194140.46  & -& 156540.49
\\ 
& mPLUG-Owl3 (7B) \citep{ye2024mplug} & 7060.15 & - &1405.26  & 42360.92  &- & 8523.82
\\ 
& Qwen2.5-VL (7B) \citep{bai2025qwen2} & 7910.18 & - & 2951.53 & 47461.08 & -&17873.64 
\\
& InternVL2.5 (8B) \citep{chen2024expanding}  & 4660.42 & - & 1198.04 & 27962.51 &- & 7063.02
\\
& DeepSeekVL (7B) \citep{lu2024deepseek}  & 32088.46 & - & 1748.68 & 192530.77 &- & 7497.00
\\ \midrule
& ResNet50 \citep{he2016deep} & 4.13 & 44.41 & 8.87 & 24.80 & 115.69 & 10.61 
\\
& ViT-B \citep{dosovitskiy2020image} & 16.86 & 80.38  & 10.66 &101.18  &352.18& 18.26
\\
& SwinT-B \citep{liu2021swin}& 15.17  & 103.05 & 30.92 & 91.02  & 414.43 & 31.29 
\\
Fine-tune & DINO v2 \citep{oquab2023dinov2}& 21.97 & 105.69 & 11.80 & 131.78 & 469.30 & 23.84 
\\
&ImageReward \citep{xu2024imagereward} & 66.88 & 283.23  & 28.80 & 401.26 & 996.10 & 107.73
\\
&HyperScore \citep{zhang2025hyperscore} & 112.09 & 215.08 & 52.80  & 168.45 & 521.93 & 55.94
\\ \midrule
proposed  &\textbf{Rank2Score} & 177.11  & 263.09 & 59.82  & 194.40 & 569.41 & 70.78
\\ \bottomrule
\end{tabular}}
\vspace{-0.4cm}
\end{table*}

\textbf{Experimental Environment.} 
To facilitate accurate evaluation, we develop an interactive annotation platform using the Three.js library \citep{threejs}. The interface allows participants to freely manipulate the camera using mouse controls for continuous viewpoint adjustment. The background is uniformly set to gray to reduce visual distractions. In addition to fully textured 3D meshes, we provide auxiliary normal-shaded views without texture, helping annotators better assess the underlying geometric structure. As shown in Fig. \ref{fig:environment_platform}, the platform enables users to navigate between samples and assign scores across 12 sub-dimensions. The subjective experiment is conducted on 27-inch AOC Q2790PQ monitors with a resolution of 2560×1440 in an indoor laboratory environment under standard lighting conditions.

\textbf{Quality Annotation.} Following prior works \citep{wu2024gpt4v,huang2023t2i,sun2025t2v,han2024evalmuse}, we employed three trained annotators to ensure annotation consistency. Each annotator independently evaluated the full set of 3,600 samples across 12 dimensions to maintain annotation robustness. To mitigate fatigue, we followed the ITU-R BT.500 guideline~\citep{bt500} and required annotators to take a break after every 30 minutes of evaluation. For samples with substantial disagreement among the three annotators (i.e., a score range of $\geq 2$), we invited three additional annotators to re-evaluate them. The re-annotation process continued until the scores fall within an acceptable agreement threshold. To avoid potential bias caused by under-specified prompts, we adopt a controlled scoring strategy. For \textit{Single}, where interaction is not applicable, the \textit{Interaction Alignment} is assigned a score of ``0'' and excluded from subsequent training and evaluation. For \textit{Multiple}, if attribute-level descriptions are absent, the \textit{Attribute Alignment} is also assigned a score of ``0" and excluded accordingly. This design ensures that each evaluation dimension is activated only when the relevant information is explicitly provided in the prompt, thereby preventing inflated scores caused by under-specification.

\section{More Details on LLM-based Evaluator}\label{app:LLM-based_evaluator}

\begin{figure*}[t]
  \centering
   \includegraphics[width=0.98\linewidth]{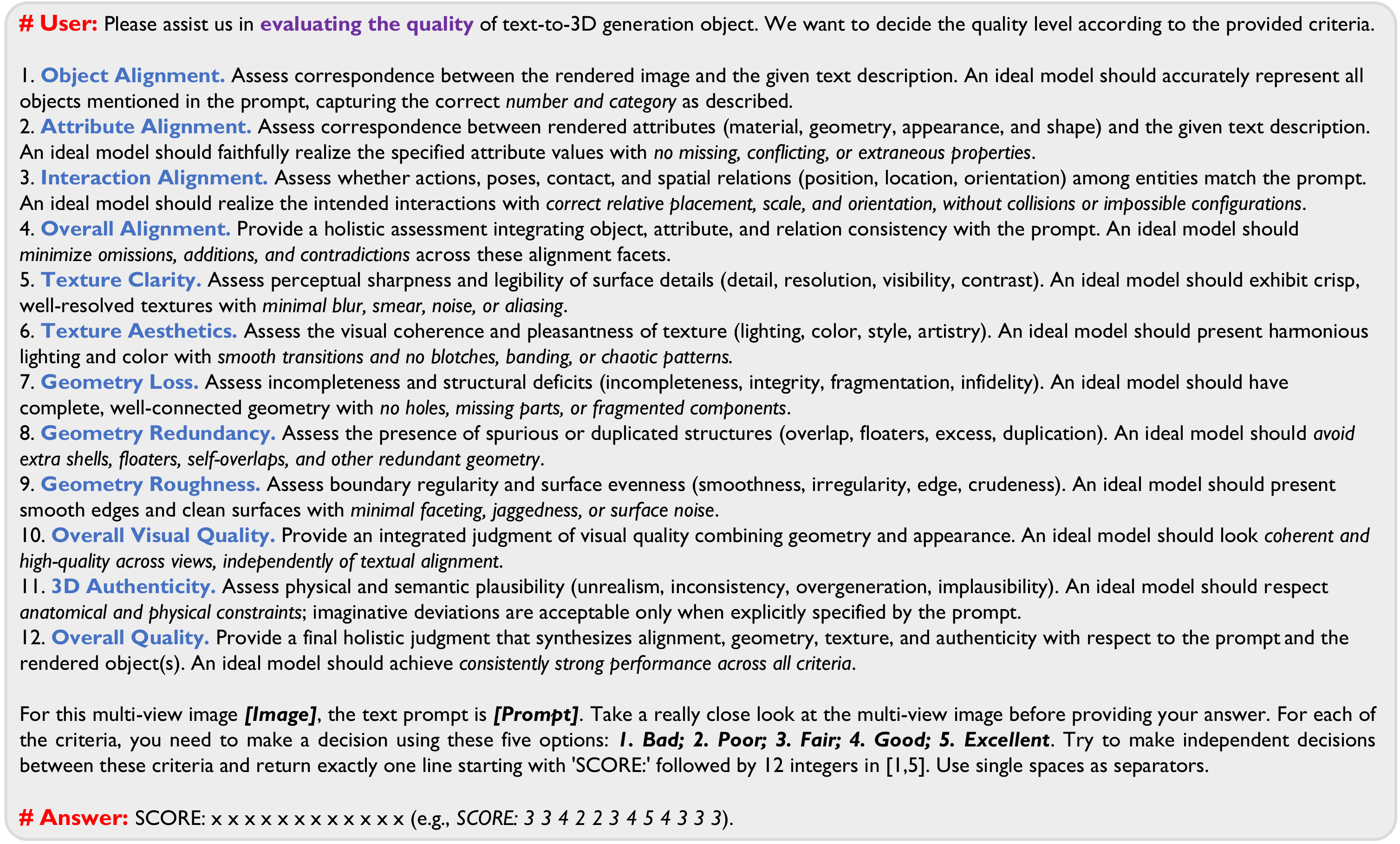}
   \caption{An example of instructions used to guide the LLM-based evaluator.}
   \label{fig:intruction_llm}
\vspace{-0.3cm}
\end{figure*}

To employ LLMs for evaluating the quality of generated 3D objects, it is necessary to provide explicit task instructions. Following the GPTEval3D \citep{wu2024gpt4v} paradigm, we first define the evaluation objective and clarify the criteria with precise descriptions. The LLM is then guided to carefully examine both the multi-view images and the associated textual prompt. Finally, we enforce a strict output format to ensure reproducibility and reliable parsing (see examples in Fig. \ref{fig:intruction_llm}). Notably, the descriptions of evaluation criteria remain flexible and can be refined or adapted to specific task requirements.

\section{Training Cost and Inference Latency}\label{app:training_cost_and_inference_latency}

We implement our metric in PyTorch and conduct all experiments on a single NVIDIA RTX 3090 GPU. Using the training protocol and hyperparameter settings described in Sec. \ref{sec:implement_details}, training on the T23D-CompBench dataset requires approximately 14 hours. Table \ref{tab:cost_and_latency_reviewer1} reports both training cost and inference latency on the T23D-CompBench benchmark under both single-view and six-view settings.

From the table, we have two observations. First, Rank2Score has a moderate inference latency. Its six-view inference latency remains in the same order of magnitude as most lightweight and fine-tuned baselines, while being much lower than large vision-language models such as LLaVA-NeXT and Qwen2.5-VL. This suggests that Rank2Score improves evaluation performance without introducing the heavy computational burden commonly associated with large-scale models.

Second, the six-view setting does not introduce a computational cost proportional to the number of views. Compared with the single-view setting, the FLOPs and inference latency of Rank2Score only increase moderately. This is because the six projected views are organized as an additional view dimension, e.g., from $(N,C,H,W)$ to $(N,6, C,H,W)$, and processed through tensorized GPU operations rather than being forwarded sequentially view by view. Therefore, the multi-view design provides a practical trade-off: it better captures 3D consistency through 2D projections while maintaining a manageable computational cost. Compared with methods based on explicit 3D processing, Rank2Score is substantially more efficient while providing stronger evaluation performance.

\begin{figure}[t]
    \centering
    \begin{subfigure}{0.45\textwidth}
        \centering
        \includegraphics[width=\linewidth]{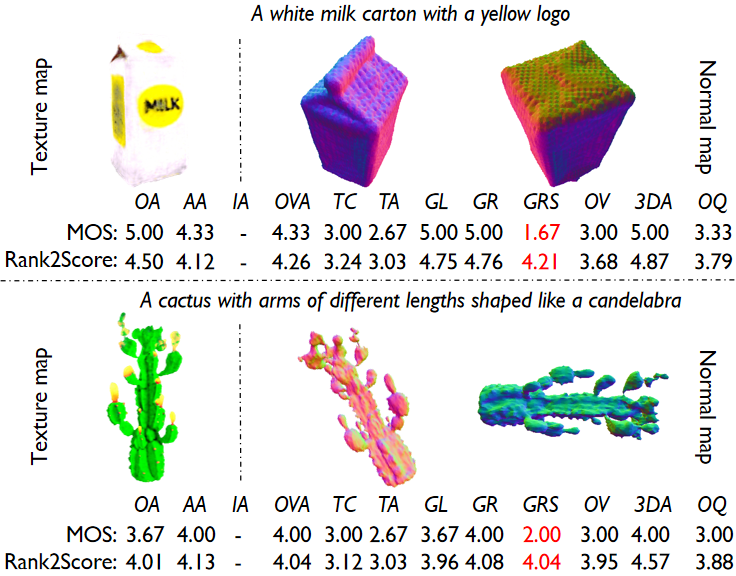}
        \subcaption{}
        \label{fig:texture_masking_1}
    \end{subfigure}
    \hspace{0.05\textwidth}
    \begin{subfigure}{0.45\textwidth}
        \centering
        \includegraphics[width=\linewidth]{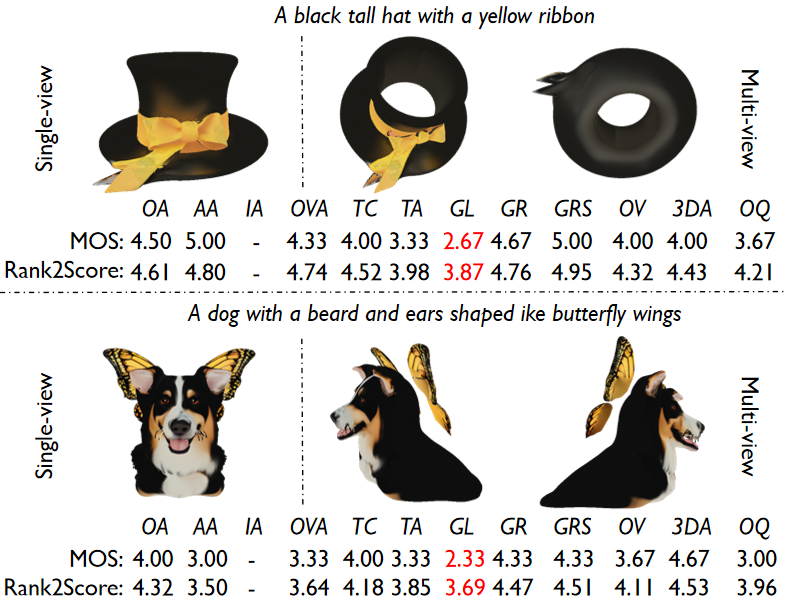}  
        \subcaption{}
        \label{fig:projection_masking_1}
    \end{subfigure}
    \caption{(a) Failure Case I: Texture Masking; (b) Failure Case II: View Omission.}
    \label{fig:review1_failure_cases}
\end{figure}

\section{Failure Case Analysis}\label{app:failure_case_analysis}

We analyze representative failure cases of Rank2Score and discuss scenarios in which 2D appearance can be misleading about the underlying 3D geometry. In our observations, these cases are relatively uncommon and mainly arise from the intrinsic limitation of projection-based evaluation. We find that they can be broadly categorized into two types: \textbf{texture masking} and \textbf{view omission}.

\textbf{Texture Masking.} Texture masking refers to cases where geometric distortions are visually obscured by surface texture. As shown in Fig.~\ref{fig:texture_masking_1}, some samples exhibit severe geometric roughness, which is clearly visible in the corresponding normal maps, but the distortion is not easily noticeable in texture renderings because the texture makes the surface appear visually plausible.

\textbf{View Omission.} View Omission refers to cases where geometric distortions are exposed only from limited viewpoints. As shown in Fig.~\ref{fig:projection_masking_1}, this issue is more likely to arise under single-view evaluation. Although multi-view alleviates this limitation, distortions visible only in a few viewpoints may be weakened in the final prediction, making the underlying geometric inconsistency harder to capture.

Overall, these failure cases mainly arise from the intrinsic limitation of projection operation. Although they cannot be completely eliminated, some designs may mitigate them in practice. Future work may further improve robustness by incorporating geometry-aware signals such as normal maps, or by directly learning geometric features from native 3D representations.

\clearpage

\end{appendices}

\bibliographystyle{spbasic}
\bibliography{bib}  

\end{document}